%% file: main.tex
\documentclass{article} %
\usepackage{iclr2027_conference,times}

\usepackage{amsmath,amssymb}
\let\papereqref\eqref
\input{math_commands.tex}

\let\eqref\papereqref
\usepackage{hyperref}
\hypersetup{colorlinks=true,linkcolor=blue,citecolor=blue,urlcolor=blue,pdfauthor={}}
\usepackage{url}
\usepackage{xcolor}
\usepackage{enumitem}
\usepackage{booktabs,graphicx}
\usepackage{placeins}
\usepackage{graphicx}
\usepackage{subcaption}
\usepackage{wrapfig}

\title{AudioLift: Audio-Driven Single Image to Full-Head Gaussian Avatar via Multi-view Video Generation}

\author{
\textbf{Zhuoyuan Jiang}$^{1}$ \enspace
\textbf{Weijie Lyu}$^{2}$ \enspace
\textbf{Xiaoyuan Wang}$^{3}$ \enspace
\textbf{Tiancheng Shen}$^{2}$ \enspace
\textbf{Ming-Hsuan Yang}$^{2}$ \\[0.5em]
$^{1}$Yale University \enspace
$^{2}$University of California, Merced \enspace 
$^{3}$Carnegie Mellon University
}

\newcommand{\ourmethod}{\textit{AudioLift}}

\iclrfinalcopy %

\begin{document}

\maketitle
\lhead{}

\begin{figure}[!htbp]
\centering
\vspace{-8mm}
\includegraphics[width=0.98\linewidth]{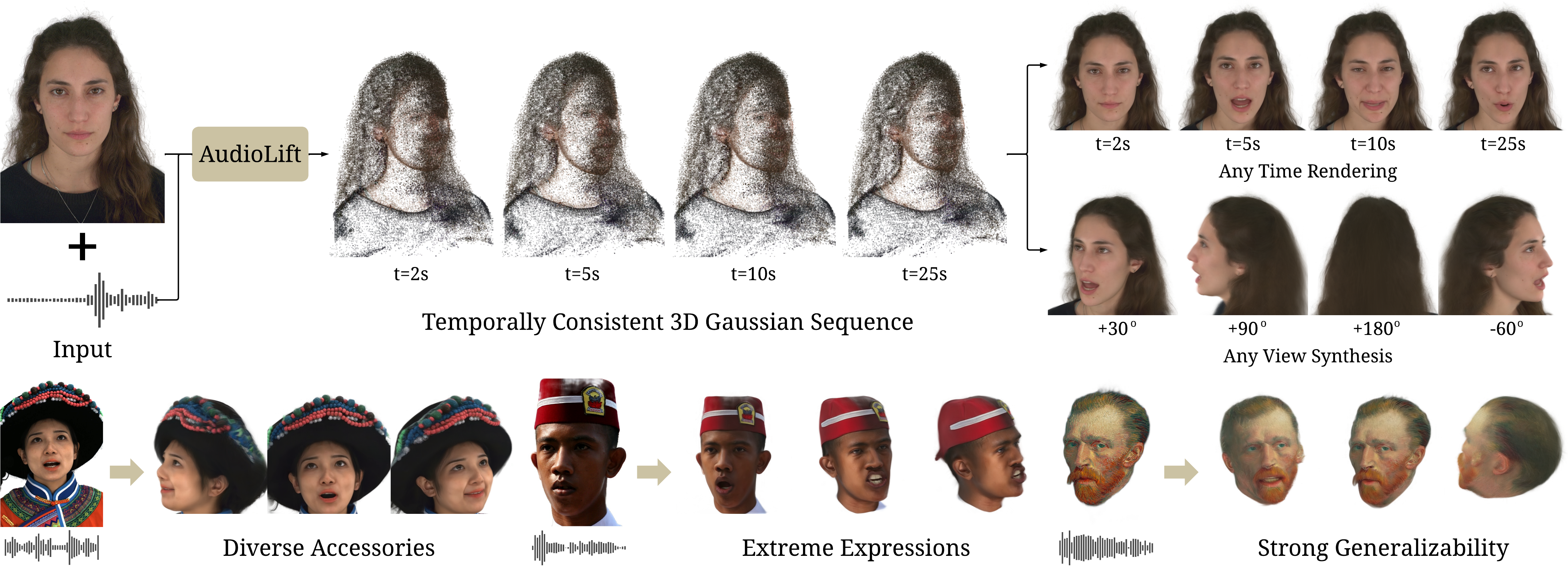}
\caption{\ourmethod{} generates a temporally coherent sequence of 3D Gaussian talking avatars from a single portrait and audio. An audio-conditioned video model jointly generates synchronized views, which a robust Gaussian reconstructor lifts into 3D. The resulting avatars support rendering from arbitrary viewpoints at any time while preserving identity and accessories, capturing expressive motion, and generalizing to diverse in-the-wild portraits.}
\label{fig:teaser}
\end{figure}

\input{SEC/0_abstract}
\input{SEC/1_intro}
\input{SEC/2_related_work}

\input{SEC/3_method}
\input{SEC/4_exp}

\input{SEC/5_conclusion}

\FloatBarrier
\clearpage

\suppressfloats[t]

\subsection*{AI use statement}

In this work, we used generative AI tools for implementation assistance on author-specified methods, translation of author-provided technical explanations into English, and feedback during discussions of evaluation procedures and result interpretation. The authors made the final experimental decisions and scientific judgments. We have not used generative AI tools for originating the core method or generating research datasets, and formal proof-related tasks are not applicable to this work. Additionally, we used generative AI tools for manuscript drafting and language refinement, table and LaTeX formatting, and identifying potentially relevant literature. Writing assistance helped organize and express author-provided technical content and findings, while literature suggestions served as candidates for subsequent review. We have reviewed all AI-assisted work. Specifically, AI-assisted code was manually inspected, suggested references were individually examined for relevance before inclusion, and AI-assisted text was checked and revised to accurately reflect the methods and findings. We take responsibility for the final content of this work, including text, claims or artifacts produced with the aid of generative AI.

\subsection*{Ethics statement}

Our work is intended for research on expressive 3D avatars and applications such as virtual communication and digital character animation. However, generating talking avatars from a single portrait and audio could enable unauthorized impersonation, fabricated speech, and misleading media. We advocate obtaining appropriate permissions for the use of personal likenesses and audio, clearly identifying generated content as synthetic, and avoiding deceptive or harmful applications. Our experiments use existing research datasets, and the sources, licenses, and attribution of additional in-the-wild assets are documented in the Appendix. Generated examples should not be interpreted as authentic recordings or statements by the depicted individuals.

\subsection*{Reproducibility statement}

Our model architecture and training objectives are described in Section~\ref{sec:method}, with additional implementation details in Appendices~\ref{app:stage1_implementation} and~\ref{app:stage2_implementation}. Appendix~\ref{app:training_recipe} specifies model initialization, training schedules, optimizer settings, random seeds, and inference configurations. Appendix~\ref{app:dataset_baselines} documents dataset construction, identity splits, preprocessing, and baseline generation. Appendix~\ref{app:metric_psnr_ssim} details the evaluation cohorts, metric implementations, aggregation rules, and ablation protocols. Together, these descriptions provide the information needed to reproduce our training and evaluation procedures.

\bibliography{iclr2027_conference}
\bibliographystyle{iclr2027_conference}

\newpage
\appendix
\input{SEC/A_inthewild}
\input{SEC/A_metrics}
\input{SEC/A_stage1_implementation}

\input{SEC/A_stage2_implementation}
\input{SEC/A_training_recipe}
\input{SEC/A_dataset_baselines}

\end{document}

%% file: math_commands.tex
\usepackage{amsmath,amsfonts,bm}

\def\eqref#1{equation~\ref{#1}}

\def\1{\bm{1}}

\DeclareMathAlphabet{\mathsfit}{\encodingdefault}{\sfdefault}{m}{sl}
\SetMathAlphabet{\mathsfit}{bold}{\encodingdefault}{\sfdefault}{bx}{n}

%% file: SEC/0_abstract.tex
\begin{abstract}
Existing 3D talking avatar methods mainly rely on parametric face models including 3DMM or FLAME, whose compact motion representations limit their ability to synthesize extreme expressions and to model accessories such as headwear.
We introduce \ourmethod{}, a two-stage framework that uses generated multi-view videos as the intermediate representation for audio-driven avatar generation without parametric face models or per-instance optimization.
Our framework combines a synchronized multi-view video generation stage with an inconsistency-robust Gaussian reconstruction stage to recover coherent full-head geometry from a single portrait image.
Extensive experiments demonstrate that \ourmethod{} produces 3D talking avatars capable of capturing more extreme facial expressions than existing methods, with improved geometric completeness and strong multi-view consistency across the full $360^\circ$ viewing range. {\urlstyle{same}\mbox{Project page: \url{https://zhuoyuanjiang.github.io/audiolift-page/}}}

\end{abstract}

%% file: SEC/1_intro.tex
\section{Introduction}
\label{sec:intro}
Audio-driven talking avatars reproduce a person’s appearance and animate facial motion from speech. They have attracted considerable attention in computer vision and graphics because of their applications in virtual communication, games, and augmented and virtual reality. Unlike 2D portrait animation, 3D talking avatars can be rendered from arbitrary viewpoints, allowing the camera to move around a speaking character.

Creating a 3D talking avatar typically requires monocular video or multi-view captures of the target person, followed by optimization for each new identity~\citep{guo2021adnerf,peng2024synctalk,aneja2025gaussianspeech}.
We instead study a more constrained setting of constructing a high-quality full-head talking avatar from only a single portrait and an audio sequence, without per-instance optimization. This would allow avatars to be created from casually captured photos.
The task presents two challenges. First, a single image leaves the geometry of the full head, especially its unseen regions, ambiguous, making consistent rendering from novel viewpoints difficult.
Second, generating realistic facial motion from audio requires mapping a one-dimensional signal to complex 3D facial dynamics while preserving temporal coherence.

A common approach first reconstructs a canonical 3D avatar, then maps audio to compact motion coefficients, often based on 3DMM~\citep{blanz1999morphable} or FLAME~\citep{li2017flame}, that deform the avatar~\citep{ye2024real3dportrait,kim2024nerffacespeech,gong2025mggtalk,he2025lam}.
While these parametric models provide useful facial structure and coordinated deformation, their low-dimensional parametric space struggles to capture subtle facial dynamics and extreme emotions, and movements of hair or accessories. To address this limitation of expressivity, recent methods try to add extra modules on top of the parametric model, such as free-form per-Gaussian deformation, to recover the dynamics it misses~\citep{xu2025vasa3d,taubner2025mvp4d,taubner2025cap4d}. However, the need for these additional deformation modules motivates us to rethink the intermediate medium for driving facial motion: could we shift the intermediate medium from a restrictive parametric representation like FLAME to a richer visual medium?

Recent audio-driven video models offer another route. Trained on large collections of talking-face videos, they can directly synthesize detailed, time-varying facial appearance from a portrait and audio, including mouth shapes and expression-dependent skin detail~\citep{tian2024emo,xu2024vasa1,cui2024hallo3,tu2025stableavatar}.
Since the generated frames already encode dense facial motion, we can bypass parametric coefficients entirely. The complex audio-to-motion synthesis is thus offloaded to the video model, while the 3D stage reconstructs a representation that explains the generated frames. This transforms 4D avatar generation from a deformation problem into a pure reconstruction task.

In this work, we present {\ourmethod}, a framework for generating a full-head 3D talking avatar from a single image and an audio sequence input. Inspired by recent advances in single-image-to-3D object generation~\citep{wonder3d, era3d, shi2023zero123pp} which generate multi-view images for lifting an image to 3D, our approach uses an audio-conditioned 2D video model ~\citep{tu2025stableavatar} built on Wan ~\citep{wan2025} to generate multi-view videos of the talking head. A feed-forward Gaussian reconstructor then lifts the generated views at each time step into a 3D Gaussian representation that can be rendered from arbitrary viewpoints~\citep{zhang2024gslrm,lyu2025facelift}.

This approach poses two challenges. First, the generated 2D videos must maintain multi-view consistency, depicting the same audio-driven performance across all viewpoints at every time step.
Since independently generating videos for individual views does not inherently guarantee cross-view consistency because of the stochastic nature of the diffusion process, we arrange videos from four oblique viewpoints in a $2 \times 2$ canvas and generate them jointly with a frontal view. In this layout, the model’s internal self-attention naturally facilitates cross-view interaction, thereby promoting appearance consistency across viewpoints.
Second, obtaining a high-quality full-head 3D representation from imperfect generated views remains challenging. Minor cross-view inconsistencies can produce floating geometry and duplicated surfaces. We address these issues by training the reconstructor with view-aware corruption and structural priors to make it robust to imperfect inputs and capable of completing unseen parts of the head. The result is a clean, compact Gaussian avatar that supports full \(360^\circ\) free-view rendering.

The main contributions are:
\begin{itemize}[leftmargin=*,nosep]
\item We introduce {\ourmethod}, a generalizable framework that creates a full-head 3D Gaussian talking avatar from a single portrait and audio, including vivid and extreme expressions.
\item We generate four audio-driven oblique views jointly in a $2 \times 2$ video grid with a frontal view, using the video model’s self-attention to promote cross-view consistency.
\item We develop a Gaussian head reconstructor that handles inconsistent views through view-aware corruption and active-volume regularization, and completes unseen rear regions through conservative rear-prior distillation.
\item Experiments on benchmark datasets and in-the-wild examples show improved identity preservation, expression quality, and novel-view fidelity over existing methods.
\end{itemize}

%% file: SEC/2_related_work.tex
\section{Related Work}
\label{sec:related}

\textbf{Audio-driven 2D talking heads.}
Early methods animate a portrait using lip synchronization, 3DMM-based warping, or diffusion conditioned on facial landmarks~\citep{prajwal2020wav2lip,zhang2023sadtalker,wei2024aniportrait,chen2024echomimic}. More recent video diffusion models generate a talking-head video directly from a portrait and audio~\citep{tian2024emo,cui2024hallo2,cui2024hallo3,lin2025omnihuman,gan2025omniavatar,tu2025stableavatar,longcat2026avatar}. By modeling appearance and motion jointly, they can capture fine facial dynamics as well as motion of hair and headwear. Their single-view outputs, however, provide no explicit 3D representation for rendering from new viewpoints. We adapt an audio-driven video model~\citep{tu2025stableavatar,wan2025} to generate synchronized views, then reconstruct a 3D Gaussian avatar from them.

\textbf{3D talking avatars and parametric head templates.}
Early methods fit a radiance field~\citep{guo2021adnerf,li2023ernerf,peng2024synctalk} or Gaussian avatar~\citep{cho2024gaussiantalker,li2024talkinggaussian,aneja2025gaussianspeech} to video or multi-view captures of each person. One-shot methods generalize across identities, but often use parametric motion: Real3D-Portrait~\citep{ye2024real3dportrait}, NeRFFaceSpeech~\citep{kim2024nerffacespeech}, and MGGTalk~\citep{gong2025mggtalk} use audio-driven 3DMM motion to animate their respective tri-plane, StyleNeRF, and point-cloud representations. Other approaches augment FLAME~\citep{li2017flame} with free-form Gaussian deformation but require optimization for each person~\citep{xu2025vasa3d,taubner2025cap4d,taubner2025mvp4d}. SDTalk~\citep{jia2026sdtalk} predicts Gaussian motion from audio in one pass but still uses FLAME to complete occluded facial regions. In contrast, we generate audio-driven multi-view video and reconstruct a full-head Gaussian avatar without parametric motion or per-person optimization.

\textbf{Multi-view generation and reconstruction.}
Generating multiple views with a diffusion model and reconstructing 3D from them is an established approach to single-image or text-to-3D generation~\citep{shi2023zero123pp,li2024instant3d,gao2024cat3d,voleti2024sv3d,lyu2025facelift,lin2025kiss3dgen,edit3r,instainpaint}. To promote cross-view consistency, methods either share self-attention across view tokens~\citep{shi2024mvdream,gao2024cat3d,kant2024spad} or introduce dedicated cross-view layers~\citep{huang2024epidiff,liu2024syncdreamer,gao2025magicdrivev2,bai2025syncammaster}. Most use image generators, whose frame-by-frame application can produce temporal flicker. Recent methods therefore use video backbones to generate multi-view sequences~\citep{shao2024human4dit,taubner2025mvp4d,cheng2026flex4dhuman,jin2026fourdanyone,facecam}, but require reference video, pose, or 3DMM controls. Our first stage generates synchronized multi-view video from only a portrait and audio; our second stage reconstructs a 3D avatar from the generated frames.

%% file: SEC/3_method.tex
\section{Method}
\label{sec:method}

\subsection{Framework Overview}
\label{sec:method_overview}

Given a reference portrait $I^{\mathrm{ref}}$ and an audio waveform $a$, {\ourmethod} synthesizes a sequence of Gaussian head avatars that follow the speech and can be
rendered from arbitrary viewpoints. We decompose this task into two stages: audio-driven multi-view video generation  and inconsistency-robust Gaussian lifting. In Stage~I, an audio-conditioned diffusion model $\mathcal{D}_{\theta}$ generates $N$ synchronized videos with $T$ frames $\{V_{1:T}^1,\ldots,V_{1:T}^N\}$, each from a distinct viewpoint. This stage can be summarized as:
\begin{equation}
  \{V_{1:T}^n\}_{n=1}^{N} = \mathcal{D}_{\theta}(I^{\mathrm{ref}},a),
  \label{eq:stage1_overview}
\end{equation}
In Stage~II, a Gaussian reconstruction model $\mathcal{R}_{\phi}$ with learnable parameters $\phi$ lifts the five generated views into a shared set of 3D Gaussians ~\citep{kerbl2023gaussian} based on their calibrated cameras $\{\Pi^1,\ldots,\Pi^N\}$. At each time step $t=1,\ldots,T$, the reconstruction process of a 3D Gaussian head avatar $\mathcal{G}_t$ can be formulated as:
\begin{equation}
  \mathcal{G}_t
    = \mathcal{R}_{\phi}\!\left(\{V^n_t,\Pi^n\}_{n=1}^{N}\right),
  \label{eq:stage2_overview}
\end{equation}
where $V^n_t$ is the $t^{th}$ frame in the $n^{th}$ generated video. 
The resulting Gaussian sequence generated by {\ourmethod} achieves high temporal consistency, benefits from the video generation process, while maintaining multi-view consistency due to the explicit Gaussian representation. {\ourmethod} enables high-quality 4D rendering. Given an arbitrary timestep $\hat{t}$ and camera viewpoint $\hat{\Pi}$, we can generate the Gaussian head $\mathcal{G}_{\hat{t}}$ to get the render image via $\hat{I} = Render(\mathcal{G}_{\hat{t}}, \hat{\Pi})$.

\subsection{Stage I: Audio-Driven Multi-View Video Generation}
\label{sec:method_stage1}

Stage~I instantiates $\mathcal{D}_{\theta}$ with StableAvatar's
audio-conditioned video diffusion transformer~\citep{wan2025,tu2025stableavatar}
(Fig.~\ref{fig:two_stage_overview}(a)). It generates five synchronized videos: four oblique views in a $2\!\times\!2$ canvas $C$ and a frontal view in a separate stream $F$. To obtain consistent multi-view evidence, we first stitch four oblique
views into a $2\!\times\!2$ canvas, allowing the four views to exchange information through spatial self-attention during denoising. Starting from this four-view model, we later found that adding a frontal view improves downstream reconstruction, as shown in our ablation. We therefore retain the
four-view canvas $C$ and add a fifth frontal stream $F$.

\input{SEC/fig_two_stage_overview}

\noindent \textbf{Audio-conditioned multi-view diffusion.}
We encode the oblique-view $2\times2$ canvas stream $C$ and frontal stream $F$ separately with the same frozen VAE, obtaining latent sequences $z_c$ and $z_f$. 
During diffusion, we add noise to both sequences at the same noise level $\sigma$, yielding $z_{c,\sigma}$ and $z_{f,\sigma}$. 
To better preserve identity information throughout denoising, we separately encode the reference portrait with the same frozen VAE to obtain a clean reference latent $z_{\mathrm{ref}}$.
We prepend $z_{\mathrm{ref}}$ along the temporal dimension to the noisy canvas latent sequence, forming $\tilde{z}_c$; the frontal
latent sequence receives a zero placeholder at the same temporal position, forming $\tilde{z}_f$. Implementation
details of reference conditioning are provided in
Appendix~\ref{app:stage1_reference}.

We pass the resulting latents $\tilde{z}_c$ and $\tilde{z}_f$ through a shared patch embedding, obtaining token blocks
$P_c^t\in\mathbb{R}^{B\times N_c\times D}$ and
$P_f^t\in\mathbb{R}^{B\times N_f\times D}$ at each latent time $t$, where $N_c,N_f$ are the per-time token counts for the canvas and frontal streams respectively. To let all five views at each latent step share the audio context for that time step, we concatenate the canvas and frontal token blocks at each latent time and flatten the resulting time groups in temporal order into one transformer sequence:
\begin{equation}
  X = \left[P_c^0 \Vert P_f^0 \Vert \cdots \Vert
             P_c^{T_\ell} \Vert P_f^{T_\ell}\right]
  \in \mathbb{R}^{B\times (T_\ell+1)(N_c+N_f)\times D}
  \label{eq:stage1_frame_pack}
\end{equation}
where $\Vert$ denotes token concatenation. Each adjacent pair
$P_c^t\Vert P_f^t$ contains the five-view tokens at latent time $t$; repeating these interleaved pairs over time forms the transformer sequence. The pair at $t=0$ is the reference-conditioning group, while $t=1,\ldots,T_\ell$ index the generated-video latent groups. 
We then assign RoPE coordinates to the combined token sequence. To avoid overlapping RoPE positions across view groups, we assign the canvas coordinates $\mathrm{RoPE}_c(t,h,w)=(t,h,w)$, while shifting the frontal stream coordinates along the width axis to $\mathrm{RoPE}_f(t,h,w)=(t,h,w+W_c^{\mathrm{tok}})$, where $W_c^{\mathrm{tok}}$ is the width of the canvas token map. As such, both streams use the same temporal index for tokens from the same video time step, while their spatial coordinates remain separate.
With these coordinates, global self-attention over $X$ acts as position-aware cross-view attention, allowing all five views to exchange information
throughout denoising.

To condition the five views on the input speech, we use cross-attention to let
visual tokens at the same latent time attend only to the audio context for the corresponding time.
Specifically, a frozen Wav2Vec2 encoder ~\citep{baevski2020wav2vec} extracts speech features from $a$. An audio adapter \citep{tu2025stableavatar} then converts each local window of these speech features into an audio context for the corresponding latent video time step. In every transformer block, canvas and frontal tokens
at the same latent time attend to the same audio context through
cross-attention, encouraging synchronized motion across views. Implementation details are provided in
Appendix~\ref{app:stage1_audio}.
After the shared output head, we reshape the sequence into
per-time canvas and frontal token blocks and unpatchify each block onto its latent
grid. Once denoising is complete, we decode both streams with the frozen VAE.

\noindent \textbf{Training objective.}
We apply StableAvatar's weighted flow-matching objective to both streams,
$\mathcal{L}_{\mathrm{FM}}=\mathcal{L}_c+\mathcal{L}_f$. For each training
sample, the canvas and frontal streams share the same sampled flow time and the same randomly selected full-frame, face-focused, or lip-focused weighting mode. We fine-tune the self-attention layers, shared patch projection, and output head, while keeping the remaining VAE and conditioning modules frozen. Appendix~\ref{app:stage1_training} specifies the loss weights, inherited auxiliary term, and training scope.

\begin{figure}[t]
\centering
\includegraphics[width=\linewidth]{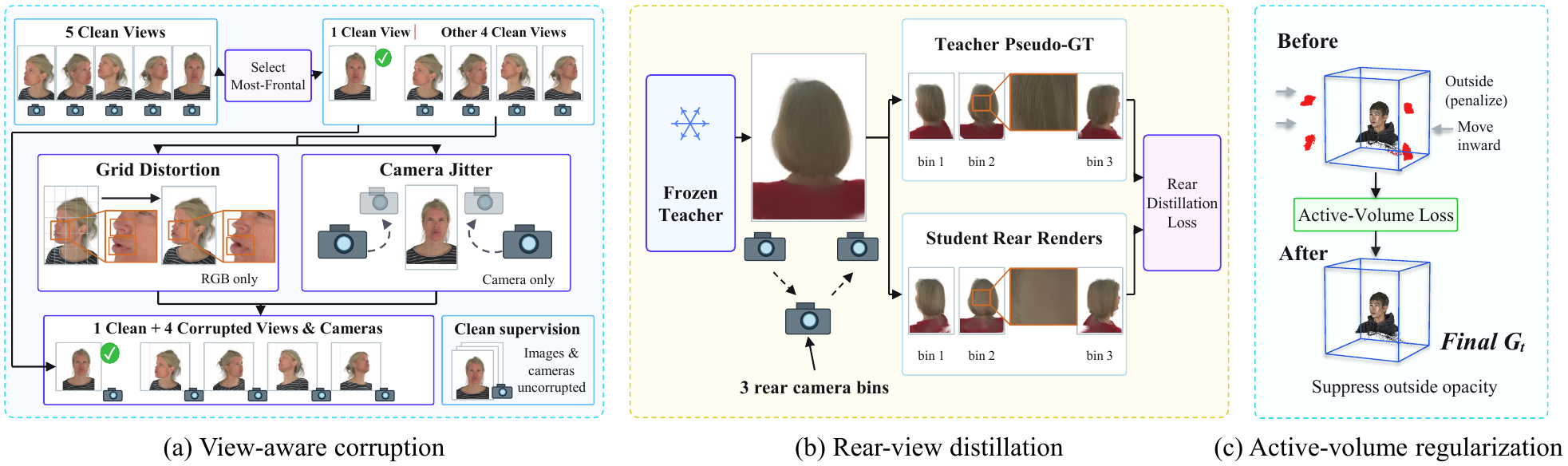}
\caption{\textbf{Stage-II training components.} (a) View-aware corruption keeps the most frontal view clean and perturbs the other four with grid distortion and camera jitter, while supervision stays uncorrupted. (b) Rear-view distillation supervises the student's rear renders against a frozen teacher in three rear camera bins. (c) Active-volume regularization reduces opacity outside the head volume and moves outlying Gaussian centers inward.}
\vspace{-2mm}
\label{fig:stage2_components}
\end{figure}

\subsection{Stage II: Inconsistency-Robust Multi-View Gaussian Lifting}
\label{sec:method_stage2}
We instantiate $\mathcal{R}_{\phi}$ in Eq.~\eqref{eq:stage2_overview} with the Gaussian reconstructor from FaceLift~\citep{lyu2025facelift}, retaining its ray-conditioned transformer and pixel-aligned Gaussian decoder. At time $t$, $\mathcal{R}_{\phi}$ predicts a 3D Gaussian head representation $\mathcal{G}_t=\{g_{t,i}\}_{i=1}^{M}$, where $M$ is the number of Gaussians and each Gaussian $g_{t,i}$ is parameterized by its 3D center, anisotropic scale, rotation,
opacity, and color.
Applying the same feed-forward lifter independently to all frames yields the ordered scenes $(\mathcal{G}_1,\ldots,\mathcal{G}_T)$, which form a 3D Gaussian avatar sequence representing our 4D avatar. Note that as the inputs of the reconstructor are generated videos showing coherent motion, the 3D Gaussian sequence naturally displays temporal consistency in practice.

\textbf{View-aware corruption.}
Although multi-view consistency of the input videos is ensured by cross-view attention in the video diffusion model, minor imperfections in facial attributes or head poses can still exist, due to the uncertainty of diffusion generation. For 3D reconstruction, these inconsistent cues can create duplicated surfaces and view-dependent floaters.
Directly fine-tuning the reconstructor with generated multi-view videos from Stage-I gave unsatisfactory results, as the generated expressions may differ from the ground truth.
We therefore train on recorded multi-view videos but apply view-aware corruption inspired by LGM ~\citep{tang2024lgm} to the input videos to mimic structured disagreement among generated views: smooth grid distortion locally warps image content while keeping its camera parameters fixed,
simulating local geometric mismatch across views; orbit-camera jitter perturbs the nominal camera pose while keeping the image fixed, simulating generated views that do not exactly follow their requested cameras.
Because the frontal view provides the strongest identity and expression cue, we keep the frontal view clean without corruptions and independently corrupt the other views. This preserves a reliable anchor while teaching the lifter to reconcile imperfect oblique evidence. All reconstruction targets remain clean.

\noindent \textbf{Active-volume regularization.}
Pixel alignment constrains each Gaussian to its source ray but leaves its depth underconstrained, allowing opacity-carrying sheets outside the head that become visible from novel views.
We therefore introduce an active-volume loss $\mathcal{L}_{\mathrm{AV}}$ that penalizes pixel-aligned Gaussians outside a canonical head volume according to both their opacity and squared distance from the volume. The opacity term makes exterior Gaussians transparent, while the distance term moves active outliers toward the boundary. The loss formulation and implementation details are provided in Appendix~\ref{app:stage2_active_volume}.

\noindent \textbf{Conservative rear distillation.}
We fine-tune the Gaussian head reconstructor with NeRSemble \citep{kirschstein2023nersemble} dataset which only captures participants from the frontal arc, so the lack of rear-view supervision can lead to poor reconstruction of the back of the head. To achieve the goal of full-head reconstruction, we distill a frozen, released FaceLift~\citep{lyu2025facelift} reconstructor whose pretraining provides a more plausible prior for rear completion. For each training scene, the teacher reconstructs the head from the fixed five clean source views. We then sample one virtual camera from each of three rear yaw ranges, $[110^\circ,157^\circ)$, $[157^\circ,203^\circ)$, and
$[203^\circ,250^\circ)$, and cache the corresponding teacher renders as three pseudo-targets. A low-weight photometric and perceptual loss is then applied to these views to encourage plausible, rather than subject-exact, completion of the unseen rear.

\noindent \textbf{Training objective and curriculum.}
Let $\mathcal{L}_{\mathrm{real}}$ and $\mathcal{L}_{\mathrm{rear}}$ denote the L2 plus $0.5$ perceptual rendering losses on real and pseudo-target views,
respectively. We optimize
\begin{equation}
  \mathcal{L}_{\mathrm{II}}
   =\mathcal{L}_{\mathrm{real}}
    +\lambda_{\mathrm{rear}}\mathcal{L}_{\mathrm{rear}}
    +\lambda_{\mathrm{AV}}\mathcal{L}_{\mathrm{AV}},
  \label{eq:stage2_total_loss}
\end{equation}
where $\lambda_{\mathrm{rear}}=0.2$ and $\lambda_{\mathrm{AV}}=0.1$ when the corresponding terms are enabled.

%% file: SEC/fig_two_stage_overview.tex
\begin{figure}[t]
  \centering
  \begin{subfigure}[t]{0.46\linewidth}
    \centering
    \includegraphics[width=\linewidth]{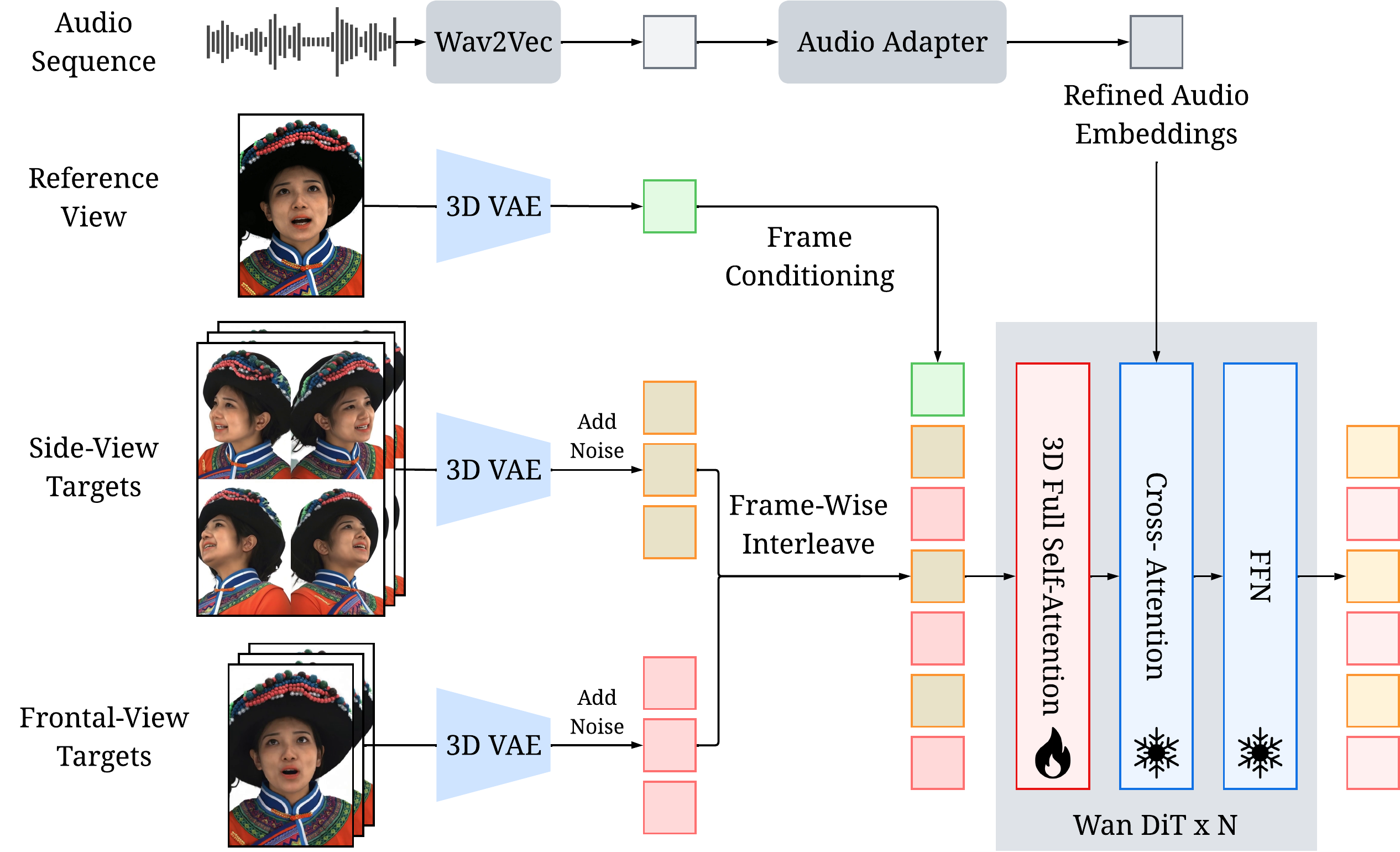}
    \subcaption{Stage I: Audio-driven multi-view generation.}
    \label{fig:stage_1}
  \end{subfigure}\hfill
  \begin{subfigure}[t]{0.52\linewidth}
    \centering
    \includegraphics[width=\linewidth]{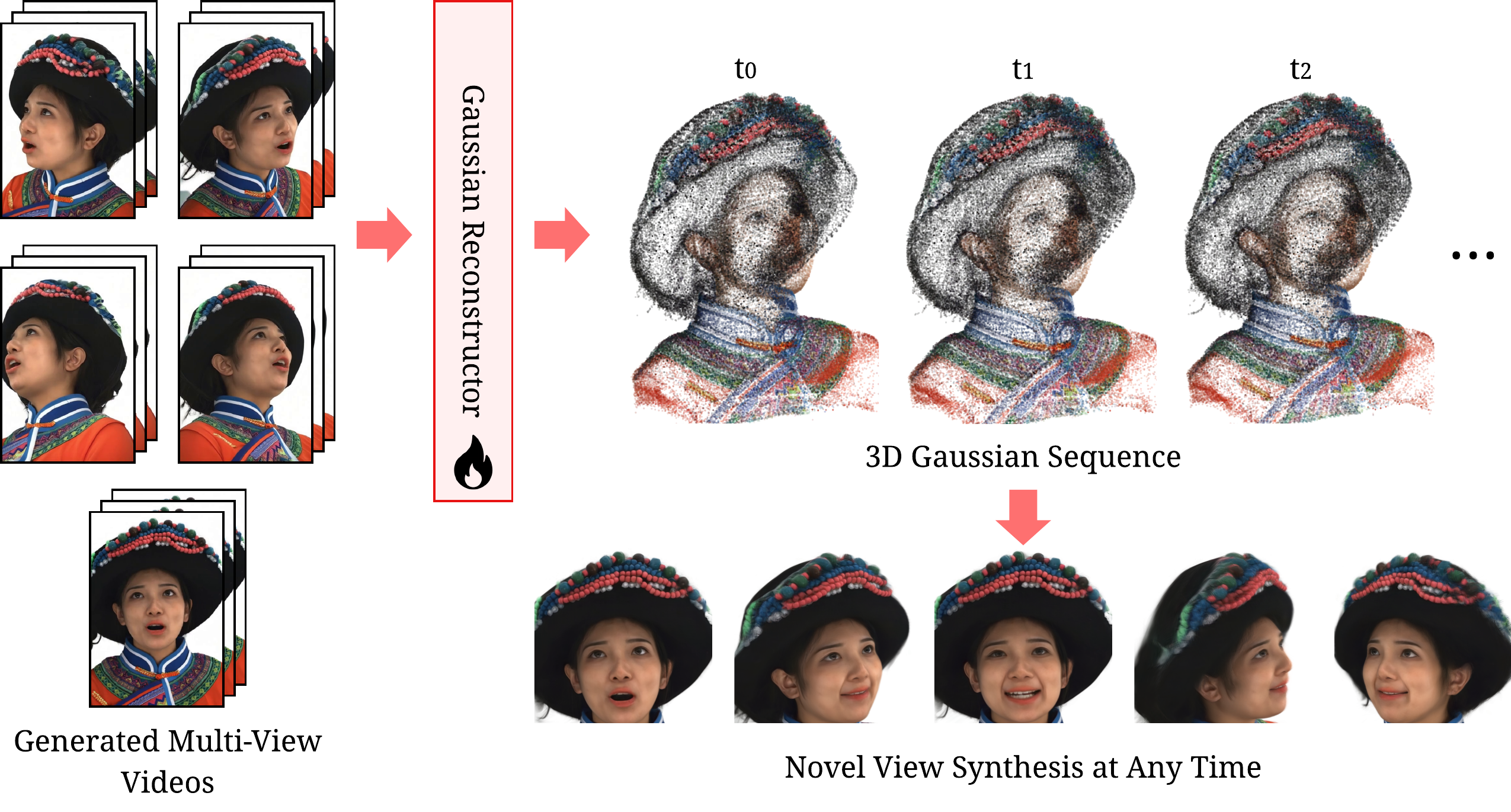}
    \subcaption{Stage II: 3D Gaussian sequence reconstruction.}
    \label{fig:stage_2}
  \end{subfigure}
  \caption{\textbf{Overview of {\ourmethod}}. (a) Five synchronized videos are generated from a single portrait and speech. The clean reference view conditions the denoiser, and the side and frontal views are interleaved into one token sequence where shared self-attention couples the views. (b) Stage II lifts generated frames at each timestamp to a 3D Gaussian avatar with a fine-tuned feed-forward reconstructor. The resulting Gaussian sequence can be rendered from any viewpoint at any time.}
\label{fig:two_stage_overview}
\end{figure}

%% file: SEC/4_exp.tex
\section{Experiments}
\label{sec:experiments}

\subsection{Experimental Setups}
\label{sec:experimental_setup}

\noindent \textbf{Datasets.}
We train on 2,641 NeRSemble ~\citep{kirschstein2023nersemble} speech sequences from 328 identities, reserving ten disjoint identities for development. Stage~I uses the full training split. Stage~II uses a fixed subset of 648 clips covering all 328 training identities, with approximately two clips per identity. We sample 16 frames per clip, yielding 10,368 synchronized multi-view scenes. For evaluation, we use 42 RenderMe-360~\citep{pan2023renderme360} clips from 21 identities with 2 clips per identity and 21 HDTF ~\citep{zhang2021hdtf} portrait-audio pairs. Quantitative evaluations on both datasets use 81-frame clips at 25 fps. Nevertheless, we support arbitrary-length inference. 
Additional in-the-wild and long-sequence demonstrations are provided on the project page. 

\noindent \textbf{Implementation details.}
Both stages use AdamW and bfloat16 on 8 NVIDIA RTX~6000 Ada
GPUs (48\,GB). Stage~I fine-tunes the 1.3B StableAvatar model~\citep{tu2025stableavatar}.
Five-view adaptation uses a global batch size of eight for 6,800 updates with a peak learning rate of $5\times10^{-4}$. It generates five $416\times608$ videos using 30 sampling steps.
Stage~II fine-tunes FaceLift~\citep{lyu2025facelift} for 7,328 updates with view-aware corruption, rear-prior distillation (weight $0.2$), and active-volume regularization (weight $0.1$).
The default global batch size is 16, with a peak learning rate of $8\times10^{-5}$.
Detailed training schedules and reproducibility settings are provided in the appendix.

\noindent \textbf{Baselines.}
We compare with Real3D-Portrait~\citep{ye2024real3dportrait},
SDTalk~\citep{jia2026sdtalk}, and
NeRFFaceSpeech~\citep{kim2024nerffacespeech} on RenderMe-360 and HDTF using
released checkpoints and the same underlying reference portraits and speech
segments. Because RenderMe-360 provides synchronized multi-view supervision,
we compare both frontal and novel-view results on it; HDTF provides only
frontal videos, so we report frontal comparisons only.

\noindent \textbf{Evaluation metrics.}
We report frontal PSNR, SSIM~\citep{wang2004ssim}, LPIPS ~\citep{zhang2018lpips}, and FID \citep{heusel2017fid} for appearance and reference-image
ArcFace ~\citep{deng2019arcface} cosine similarity (CSIM-ref) for identity.
Frontal appearance metrics use frames 1--80 after mapping every method to a
common $512\times512$ white-background scoring canvas;  Novel-view metrics use frame~0 at five unseen cameras on the same scoring canvas. Preprocessing,
metric implementations, and aggregation rules are detailed in the appendix.

\begin{table}[b]
\centering
\small
\vspace{-2mm}
\caption{End-to-end comparison on RenderMe-360. Best results are bolded.}
\vspace{-2mm}
\label{tab:renderme_end_to_end}
\begin{tabular}{@{\hskip 0.8mm}l@{\hskip 0.8mm}c@{\hskip 0.8mm}c@{\hskip 0.8mm}c@{\hskip 0.8mm}c@{\hskip 0.8mm}c@{\hskip 0.8mm}c@{\hskip 0.8mm}c@{\hskip 0.8mm}c@{\hskip 0.8mm}}
\hline
& \multicolumn{5}{c}{(a) Frontal comparison} &
  \multicolumn{3}{c}{(b) Novel-view comparison} \\
\cline{2-6}\cline{7-9}
Method & PSNR$\uparrow$ & SSIM$\uparrow$ & LPIPS$\downarrow$ & FID$\downarrow$ & CSIM-ref$\uparrow$ & PSNR$\uparrow$ & SSIM$\uparrow$ & LPIPS$\downarrow$ \\
\hline
Real3D-Portrait \citep{ye2024real3dportrait} & 17.50 & 0.700 & 0.246 & 83.64 & 0.650 & 10.82 & 0.557 & 0.491 \\
SDTalk \citep{jia2026sdtalk} & 17.63 & \textbf{0.710} & 0.232 & 77.05 & 0.645 & 10.99 & 0.571 & 0.473 \\
NeRFFaceSpeech \citep{kim2024nerffacespeech} & 11.10 & 0.505 & 0.466 & 108.08 & 0.398 & 8.51 & 0.426 & 0.622 \\
\textbf{Ours} & \textbf{18.62} & 0.678 & \textbf{0.216} & \textbf{44.53} & \textbf{0.689} & \textbf{12.27} & \textbf{0.582} & \textbf{0.445} \\
\hline
\end{tabular}
\end{table}

\subsection{Main Results}
\label{sec:end_to_end_comparison}

\noindent \textbf{Comparison on RenderMe-360.}
Table~\ref{tab:renderme_end_to_end} compares frontal-view and novel-view
performance on RenderMe-360. The novel-view evaluation tests whether the
reconstructed avatar generalizes from the frontal reference to non-frontal
viewpoints. As shown in the table, our method achieves a substantially lower
FID than all baselines (44.53 vs. 77.05--108.08), indicating that its generated
videos better match the distribution of real videos. On the novel-view camera
set, our method also achieves the best PSNR, SSIM, and LPIPS among all baselines,
demonstrating stronger generalization to non-frontal viewpoints.

\begin{table}[t]
\small
\centering
\caption{Frontal comparison on HDTF. Best results are bolded. Second-best results are underlined.}
\vspace{-2mm}
\label{tab:hdtf_end_to_end}
\begin{tabular}{lccccc}
\hline
Method & PSNR$\uparrow$ & SSIM$\uparrow$ & LPIPS$\downarrow$ & FID$\downarrow$ & CSIM-ref$\uparrow$ \\
\hline
Real3D-Portrait \citep{ye2024real3dportrait} & 13.69 & 0.650 & 0.344 & \underline{35.31} & \textbf{0.671} \\
SDTalk \citep{jia2026sdtalk} & \underline{13.87} & \underline{0.651} & \underline{0.338} & 40.53 & 0.635 \\
NeRFFaceSpeech \citep{kim2024nerffacespeech} & 10.58 & 0.561 & 0.482 & 93.72 & 0.391 \\
\textbf{Ours} & \textbf{13.98} & \textbf{0.653} & \textbf{0.335} & \textbf{29.73} & \underline{0.666} \\
\hline
\end{tabular}
\end{table}

\noindent \textbf{Comparison on HDTF.}
Table~\ref{tab:hdtf_end_to_end} reports the frontal-only comparison on HDTF. It tests generalization to in-the-wild portraits and
audio. We omit novel-view metrics because HDTF does not provide
synchronized multi-camera ground truth. Our method obtains the best PSNR, SSIM, LPIPS, and FID while achieving CSIM-ref scores very close to the best-performing baseline (0.666 vs. 0.671). These results indicate that our method produces avatars with strong perceptual quality and rendered appearances that closely match the distribution of real-world face images.

\begin{figure}[tb]
\centering
\vspace{-2mm}
\includegraphics[width=0.98\linewidth]{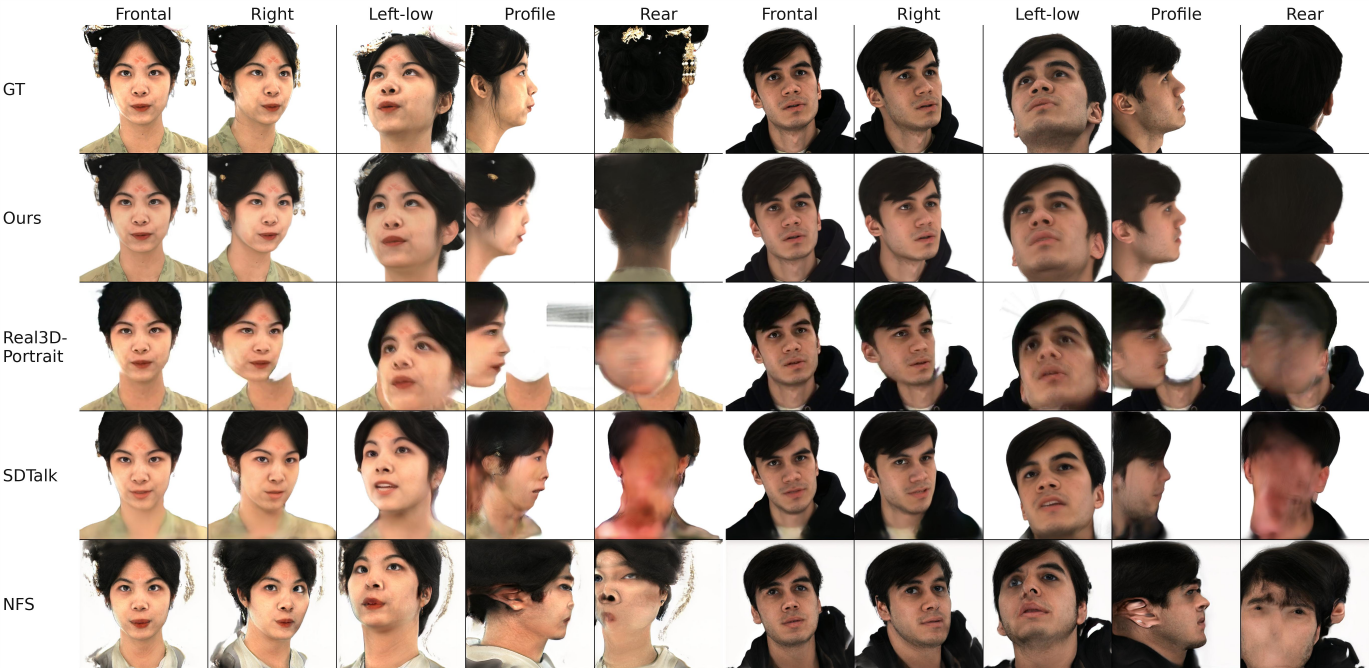}
\caption{Same-frame multi-view comparison on RenderMe-360. Rows are ground truth, {\ourmethod}, Real3D-Portrait~\citep{ye2024real3dportrait}, SDTalk~\citep{jia2026sdtalk}, and NeRFFaceSpeech~\citep{kim2024nerffacespeech}; columns move from the frontal camera through oblique and profile views to the rear. The results show that {\ourmethod} produces facial expressions closer to the ground truth. Meanwhile, {\ourmethod} also achieves better novel-view synthesis results, and faithfully maintains the head accessories from the reference image.}
\vspace{-2mm}
\label{fig:end_to_end_qualitative}
\end{figure}

\begin{figure}[tb]
\centering
\includegraphics[width=0.8\linewidth]{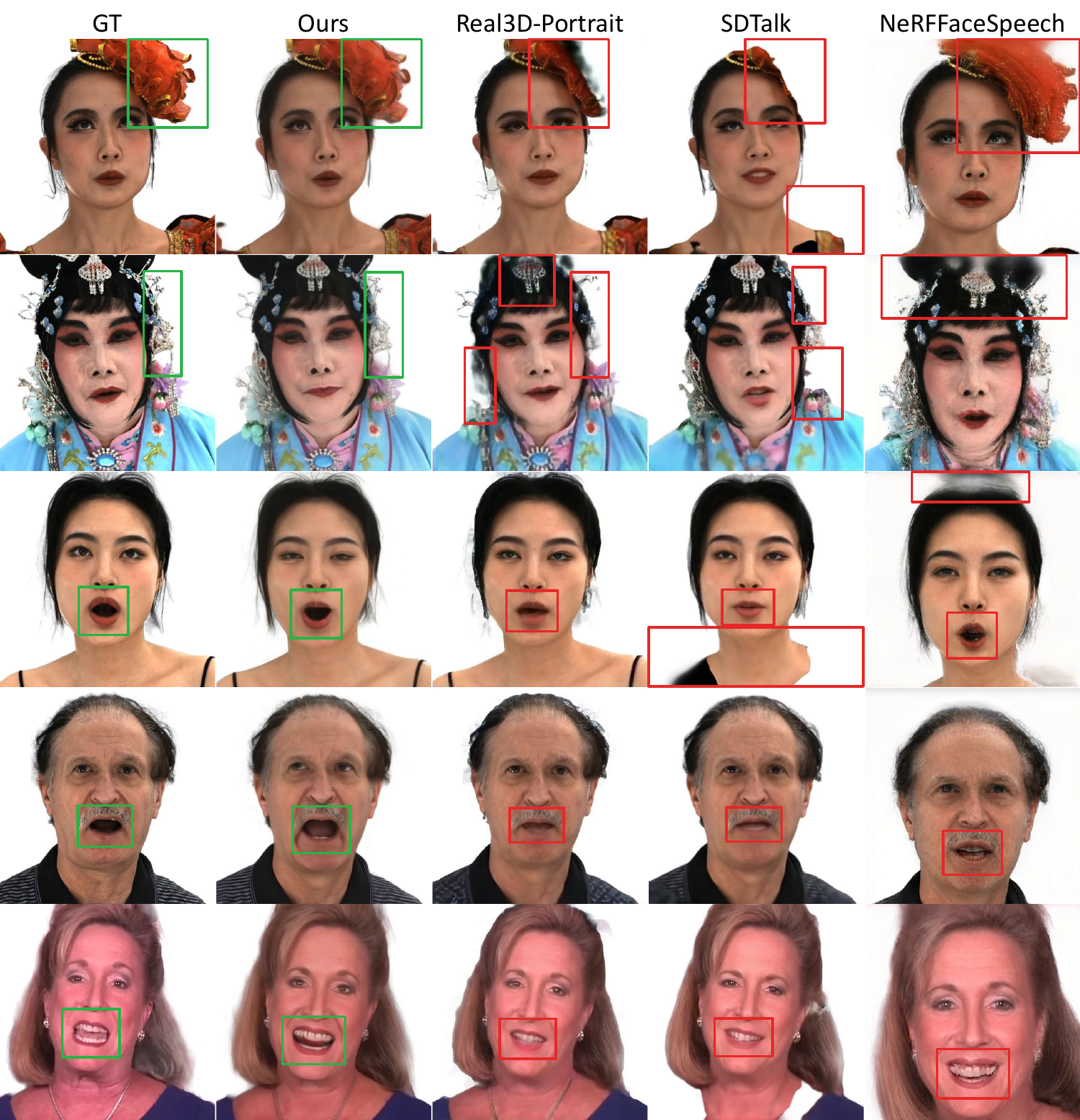}
\caption{Qualitative comparison under challenging headwear and facial expressions. The first two rows emphasize preservation of fine headwear, while the remaining rows compare expressive mouth shapes. Green boxes mark target expressions reproduced by our method; red boxes highlight missing details, boundary artifacts, or expression mismatches in the baselines.}
\label{fig:hdtf_headwear_expression}
\end{figure}

\noindent \textbf{Qualitative comparison.}
Figure~\ref{fig:end_to_end_qualitative} compares all methods' multi-view consistency and geometric completeness of reconstructed avatars on RenderMe-360. 
Specifically, Real3D-Portrait shows head-neck separation as the viewpoint changes and fails to reconstruct a plausible back of the head. SDTalk shows artifacts across side views, inconsistent facial expressions when changing viewpoint, and blurred rear appearances. NeRFFaceSpeech exhibits distorted facial structures in non-frontal views. In contrast, our method maintains a coherent head shape and face-neck structure across viewpoints and reconstructs a plausible back of the head rather than blurred artifacts. These comparisons highlight the advantage of our method in reconstructing complete and multi-view-consistent avatars beyond just the frontal view. 
Beyond multi-view consistency, the RenderMe-360 and HDTF examples in Figure~\ref{fig:hdtf_headwear_expression} highlight our method's ability to preserve accessories and capture pronounced facial expressions. Specifically, while other methods omit parts of accessories, our method preserves them well. Our method also reproduces wide mouth openings more faithfully, whereas other methods show more restricted mouth openings. Video comparisons are provided on the project page.

\subsection{Ablation Studies}
\label{sec:ablations}

\begin{figure}[t]
  \centering
  \begin{subfigure}[t]{0.42\linewidth}
    \centering
    \includegraphics[width=\linewidth]{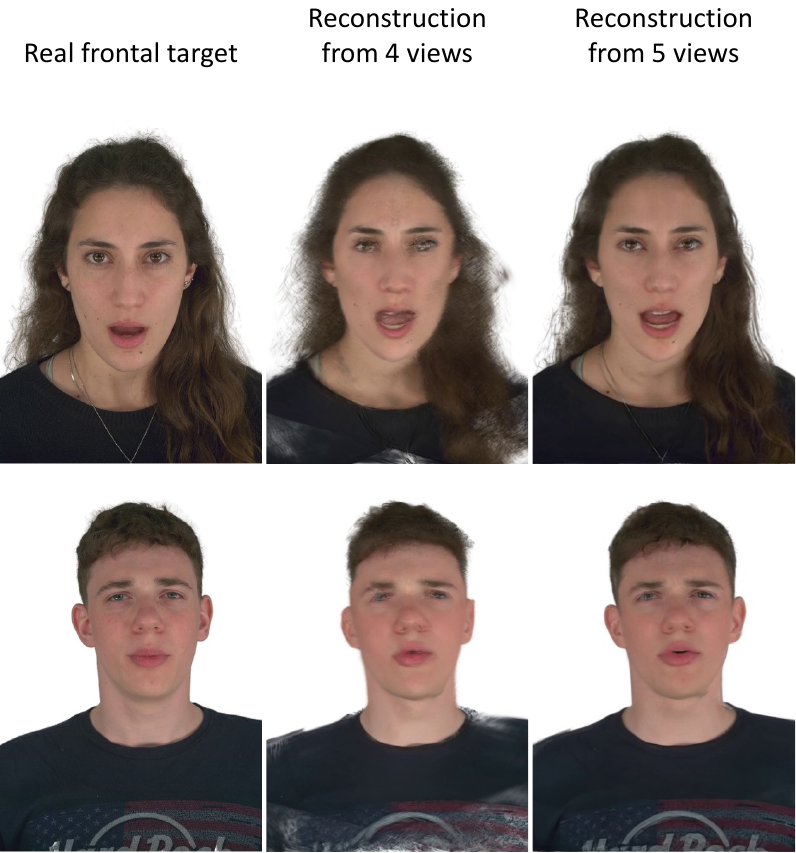}
    \subcaption{Qualitative results of ablation study on number of generated multi-view videos.}
    \label{fig:ablation_view}
  \end{subfigure}\hfill
  \begin{subfigure}[t]{0.56\linewidth}
    \centering
    \includegraphics[width=\linewidth]{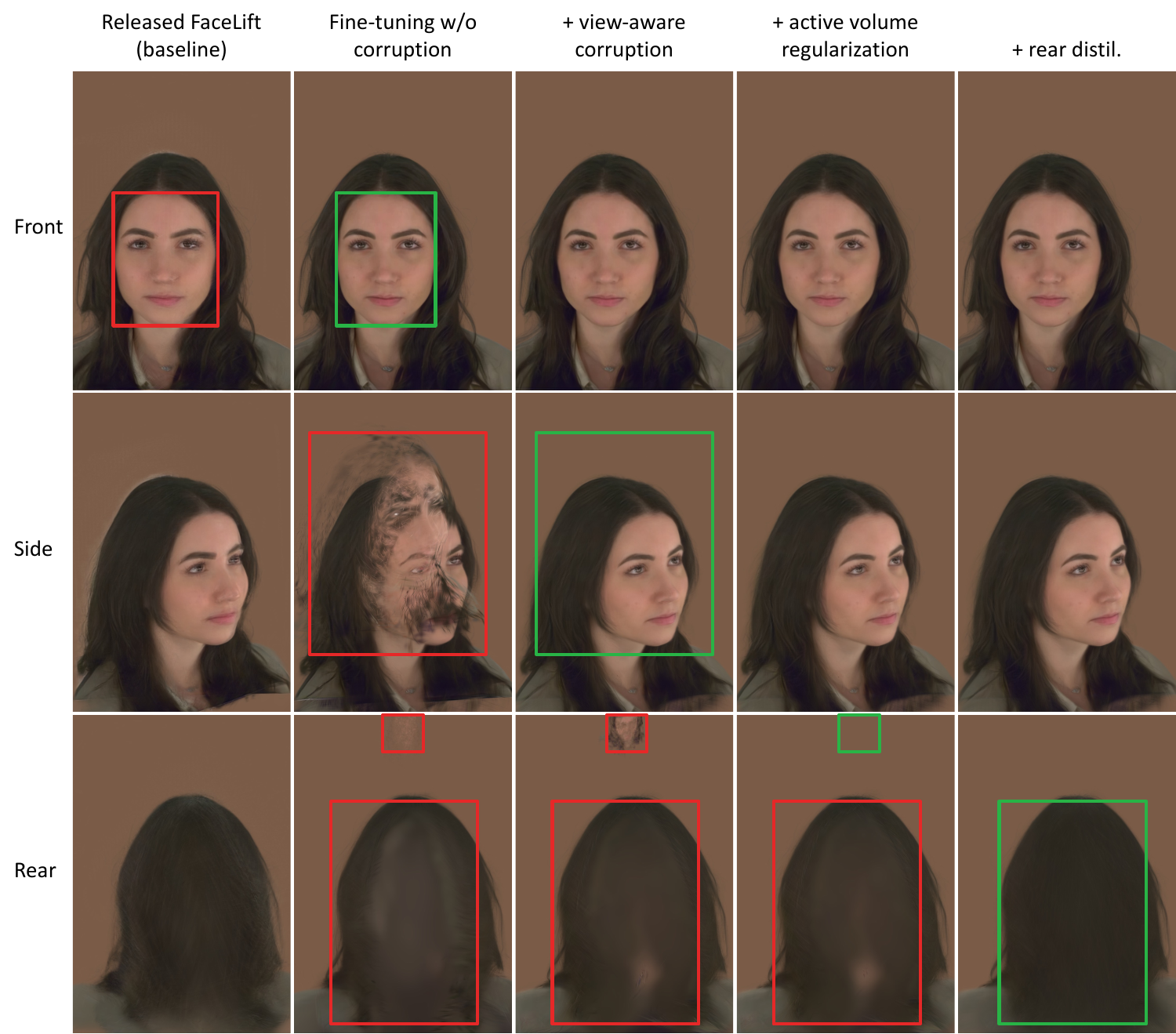}
    \subcaption{Qualitative results on reconstructor finetuning components.}
    \label{fig:ablation_ft}
  \end{subfigure}
  \caption{Ablation study results visualization. (a) With the additionally front-view videos as input, the reconstruction results demonstrate obviously fewer artifacts. (b) Training with view-aware corruption helps model learn to resolve minor view inconsistency, active volume regularization removes the outlying Gaussian floaters, while rear distillation improves the visual results of back of the head.}
\label{fig:ablations_side_by_side}
\end{figure}

\noindent \textbf{Stage I design choices.}
We evaluate the effect of adding a fifth frontal view to the downstream reconstruction inputs. Keeping the generated four-view inputs and reconstruction checkpoint fixed, we compare reconstruction with and without the additional frontal view. 
Figure~\ref{fig:ablations_side_by_side}(a) shows that the fifth view produces sharper frontal reconstructions that more closely match the ground truth.
Table~\ref{tab:stage1_four_five_views}(a) also shows substantial improvements across all three metrics, confirming that the additional frontal view improves downstream reconstruction quality.

\begin{table}[!ht]
\centering
\caption{Quantitative ablations of (a) Stage I: number of multi-view videos and (b) Stage II: effectiveness of view-aware corruption. Best results are bolded.}
\label{tab:quantitative_ablations}                                                                                                                                                       
\label{tab:stage1_four_five_views}
\label{tab:stage2_checkpoint_comparison}
\small
\textbf{(a) Fifth Frontal View}\par\smallskip
\begin{tabular}{lccc}
\toprule
Reconstruction Inputs & PSNR$\uparrow$ & SSIM$\uparrow$ & LPIPS$\downarrow$ \\
\midrule
4-view videos & 16.19 & 0.7190 & 0.2843 \\
5-view videos & \textbf{20.18} & \textbf{0.7787} & \textbf{0.1803} \\
\bottomrule
\end{tabular}
\par\medskip
\textbf{(b) View-aware Corruption}\par\smallskip
\begin{tabular}{lccc}
\toprule
Model & PSNR$\uparrow$ & SSIM$\uparrow$ & LPIPS$\downarrow$ \\
\midrule
FaceLift baseline \citep{lyu2025facelift} & 17.39 & 0.7255 & 0.2763 \\
Fine-tuned w/o corruption & 17.77 & 0.7314 & 0.2606 \\
Fine-tuned w/ corruption & \textbf{17.89} & \textbf{0.7344} & \textbf{0.2573} \\
\bottomrule
\end{tabular}
\end{table}

\noindent \textbf{Stage II components.}
Figure~\ref{fig:ablations_side_by_side} evaluates the contributions of view-aware
corruption, active-volume regularization, and rear-prior distillation.
Fine-tuning FaceLift on our data sharpens facial details. However, because the
synthesized input views are not perfectly consistent~\citep{vgc}, direct fine-tuning can
fit the observed cameras with 3D geometry that satisfies those views but does
not remain correct from other viewpoints, producing floating structures around
the head in side and rear renderings. Training with view-aware corruption
removes most of these artifacts and produces a more compact reconstruction by
teaching the model to reconcile inconsistencies across views. Nevertheless,
some opaque Gaussians
remain outside the spatial region expected to contain the head and upper body,
as visible in the last row of the corruption result. Active-volume
regularization suppresses these misplaced Gaussians and produces a cleaner,
more compact reconstruction. Finally, rear-prior distillation improves the appearance of the back of the head. Together, these ablations show that each component addresses a distinct failure mode and contributes to the final reconstruction quality. Table~\ref{tab:stage2_checkpoint_comparison}(b) provides quantitative results that further support the effectiveness of view-aware corruption.

%% file: SEC/5_conclusion.tex
\FloatBarrier
\vspace{-1mm}
\section{Conclusion}
\vspace{-1mm}
\label{sec:conclusion}
We introduce \ourmethod{}, a two-stage framework that generates a 3D talking avatar from a single portrait and audio, without a parametric face model or per-instance optimization. It jointly generates audio-driven multi-view video and reconstructs a full-head Gaussian avatar while accounting for inconsistencies among the generated views. Experiments on RenderMe-360 and HDTF, together with in-the-wild examples, show expressive facial motion, consistent novel-view rendering, and improved geometric completeness across a full $360^\circ$ viewing range.

%% file: SEC/A_inthewild.tex
\section{Additional In-the-Wild Results}
\label{app:inthewild}

This appendix collects qualitative results on in-the-wild portraits: paintings,
anime, stylized art, and photographs, driven by speech in English, Japanese,
and Swahili. Frames in Sections~\ref{app:itw_comparison}
and~\ref{app:itw_temporal} are selected by phoneme, that is, the clearest
instance of each target sound among the frames where the audio is most
certain. Alongside these selected frames, Table~\ref{tab:itw_mouth_opening}
reports an auxiliary mouth-opening measurement using automatically selected
phoneme times across all 16 clips in the same curated set.

\subsection{Comparison with Prior Methods}
\label{app:itw_comparison}

Across these examples the baselines animate the mouth far less than the audio
asks for. The gap is clearest on open vowels: where the phoneme requires a
wide opening, our method opens the mouth wide and closes it fully on the
bilabial stops \mbox{/m/}, \mbox{/b/}, and \mbox{/p/}, while the baselines
hold a narrow, nearly constant aperture throughout. Table~\ref{tab:itw_mouth_opening}
quantifies exactly this contrast over all clips.

\begin{table}[htbp]
\centering
\caption{Mouth-opening contrast on all 16 in-the-wild clips, measured as the mean mouth
opening at open vowels minus the mean opening at the bilabials
\mbox{/m/}, \mbox{/b/}, \mbox{/p/}. Phoneme times are selected automatically
within the curated set. Implementation details are in
Appendix~\ref{app:mouth_opening_contrast}.}
\label{tab:itw_mouth_opening}
\begin{tabular}{lc}
\toprule
Method & Mouth-opening contrast $\uparrow$ \\
\midrule
Ours & \textbf{0.166} \\
NeRFFaceSpeech & 0.122 \\
Real3D-Portrait & 0.084 \\
SDTalk & 0.078 \\
\bottomrule
\end{tabular}
\end{table}

\begin{figure}[htbp]
\centering
\includegraphics[width=\linewidth]{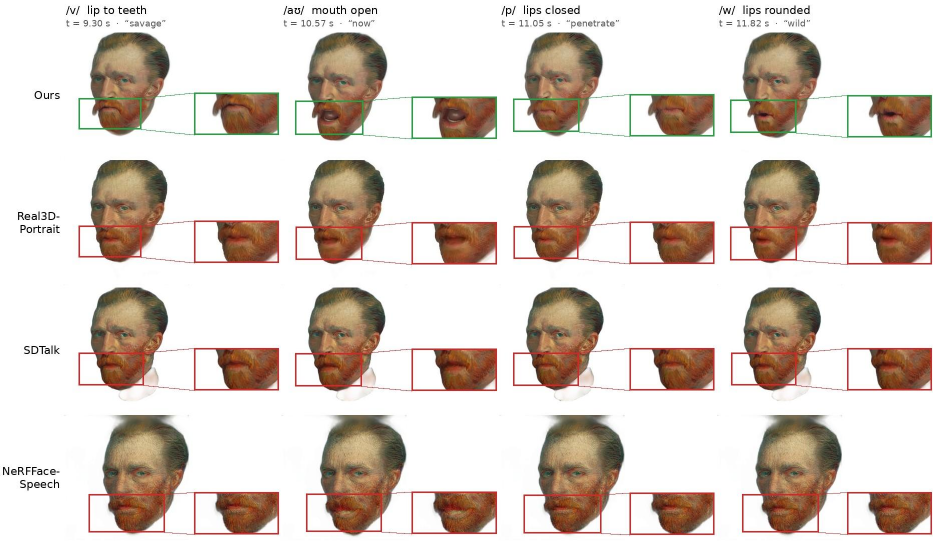}
\caption{Van Gogh self-portrait, English speech. Our method opens the mouth on
``now'', closes the lips on the \mbox{/p/} of ``penetrate'', and rounds them on
``wild''; the three baselines keep almost the same mouth shape throughout.}
\label{fig:itw_a1}
\end{figure}

\begin{figure}[htbp]
\centering
\includegraphics[width=\linewidth]{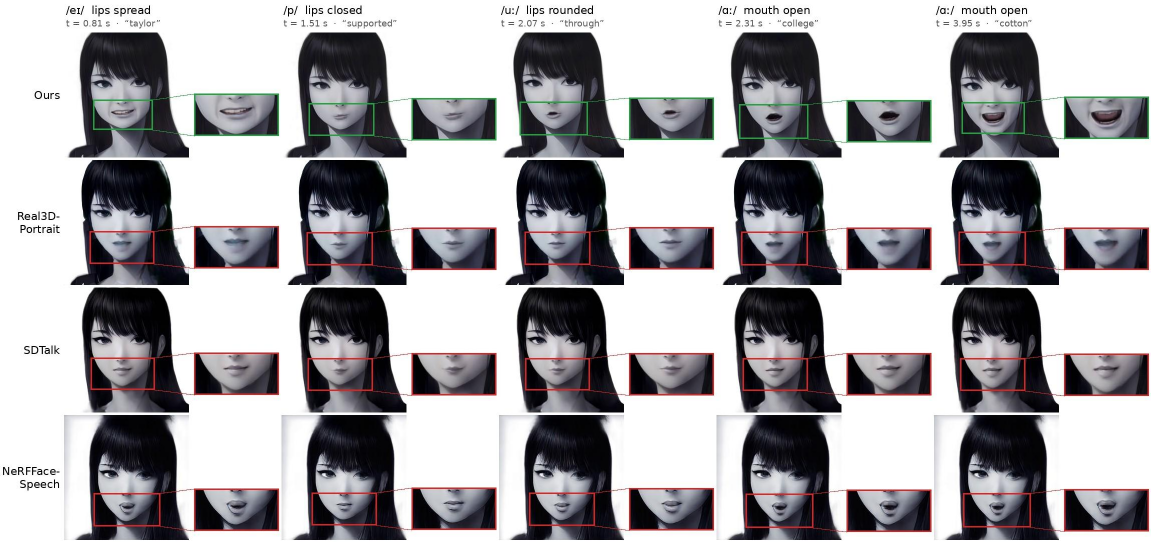}
\caption{Anime portrait, English speech. Five phonemes give five distinct
mouth shapes in our result: spread lips on ``Taylor'', closed lips on
``supported'', rounded lips on ``through'', open on ``college'', and wide open
on ``cotton''. Real3D-Portrait also shifts the colour of the drawing.}
\label{fig:itw_a2}
\end{figure}

\begin{figure}[htbp]
\centering
\includegraphics[width=\linewidth]{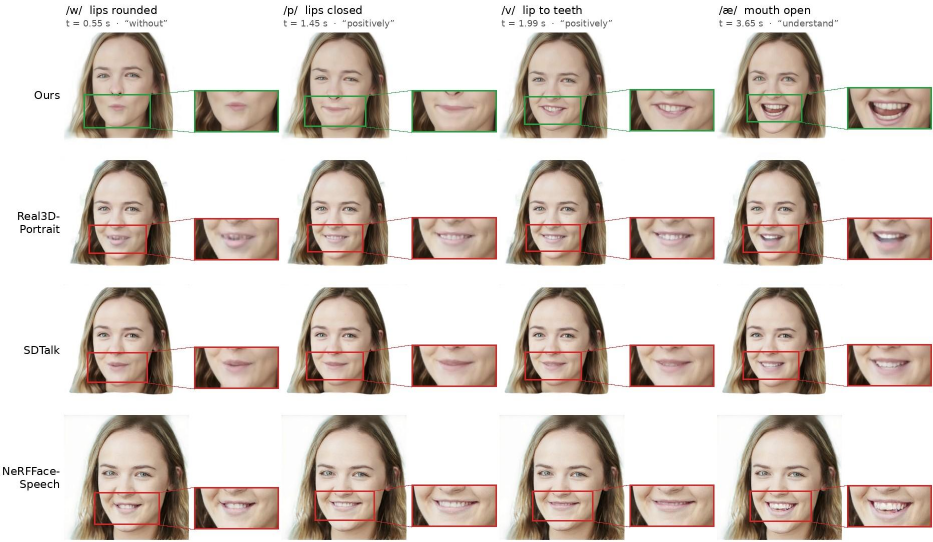}
\caption{Photographic portrait, English speech. Only our method purses the lips
for the \mbox{/w/} of ``without'', closes them for the \mbox{/p/} of
``positively'', and opens the jaw on ``understand''.}
\label{fig:itw_a3}
\end{figure}

\begin{figure}[htbp]
\centering
\includegraphics[width=\linewidth]{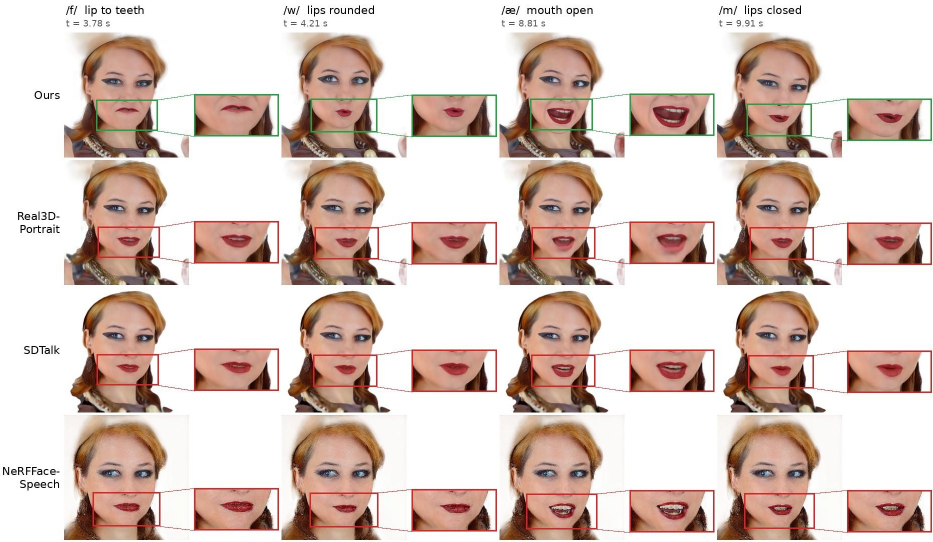}
\caption{Heavy makeup and earrings, sung English. Our method reproduces all
four target mouth shapes, while the baselines stay close to a single shape.}
\label{fig:itw_a4}
\end{figure}

\begin{figure}[htbp]
\centering
\includegraphics[width=\linewidth]{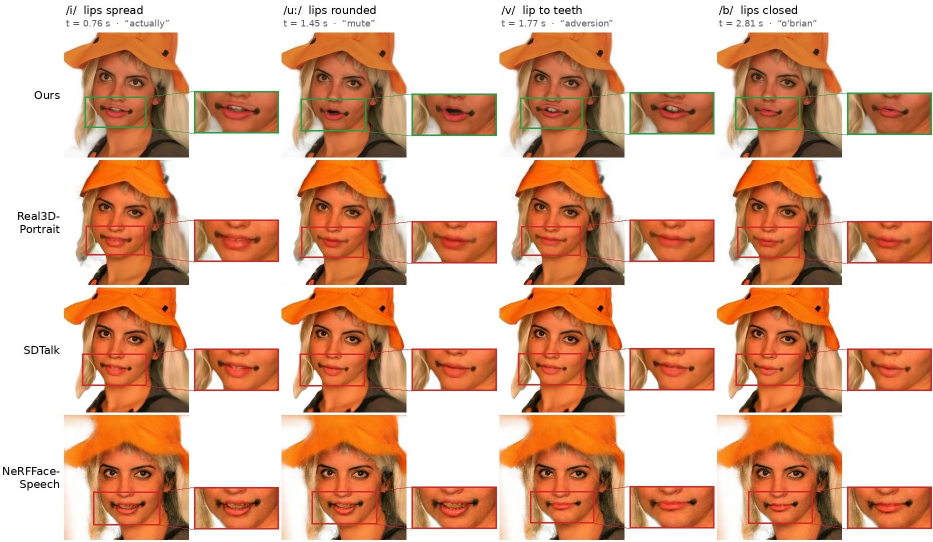}
\caption{Wide-brimmed hat and face paint, English speech. Our method presses
the upper teeth to the lower lip for \mbox{/v/} and closes the lips for
\mbox{/b/}; NeRFFaceSpeech blurs the hat.}
\label{fig:itw_a5}
\end{figure}

\begin{figure}[htbp]
\centering
\includegraphics[width=\linewidth]{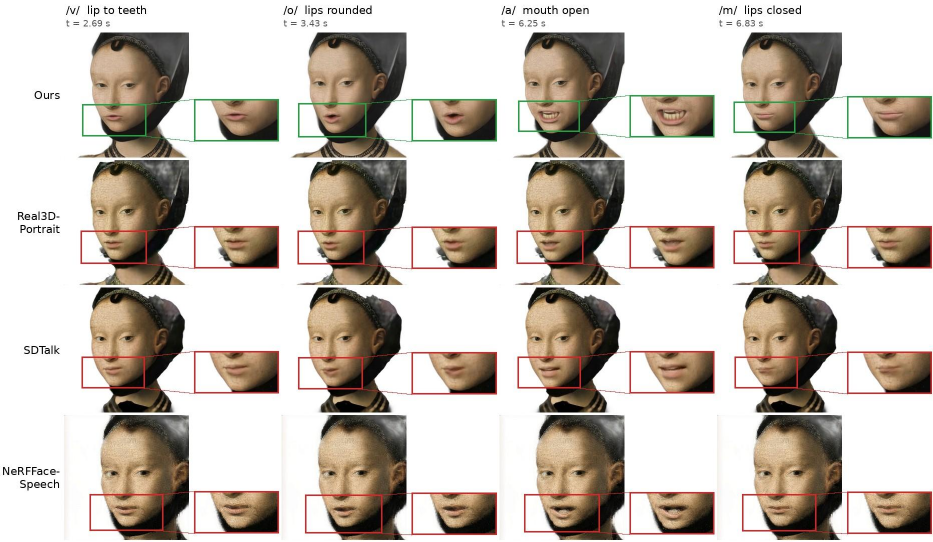}
\caption{Renaissance painting, Swahili speech. NeRFFaceSpeech shows a grid
pattern across the face.}
\label{fig:itw_a7}
\end{figure}

\begin{figure}[htbp]
\centering
\includegraphics[width=\linewidth]{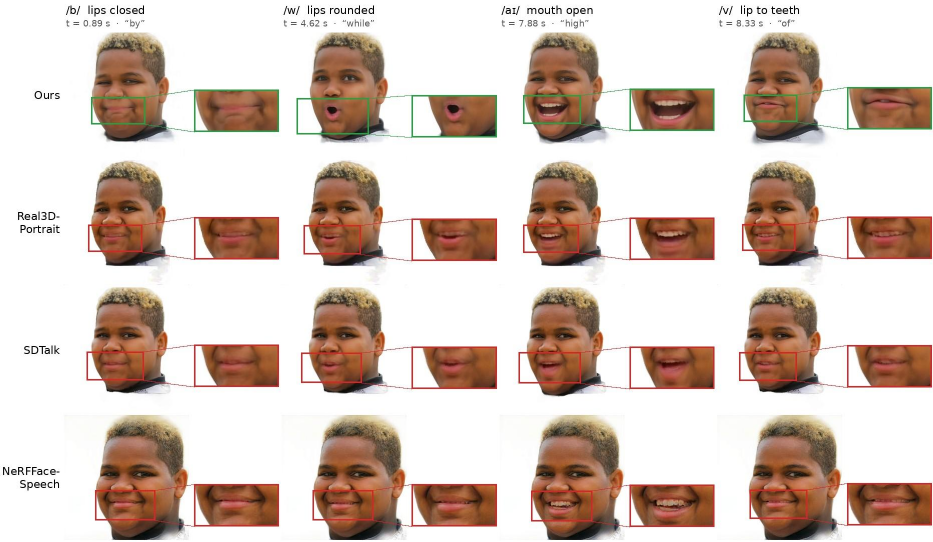}
\caption{Child portrait, English speech.}
\label{fig:itw_a8}
\end{figure}

\begin{figure}[htbp]
\centering
\includegraphics[width=\linewidth]{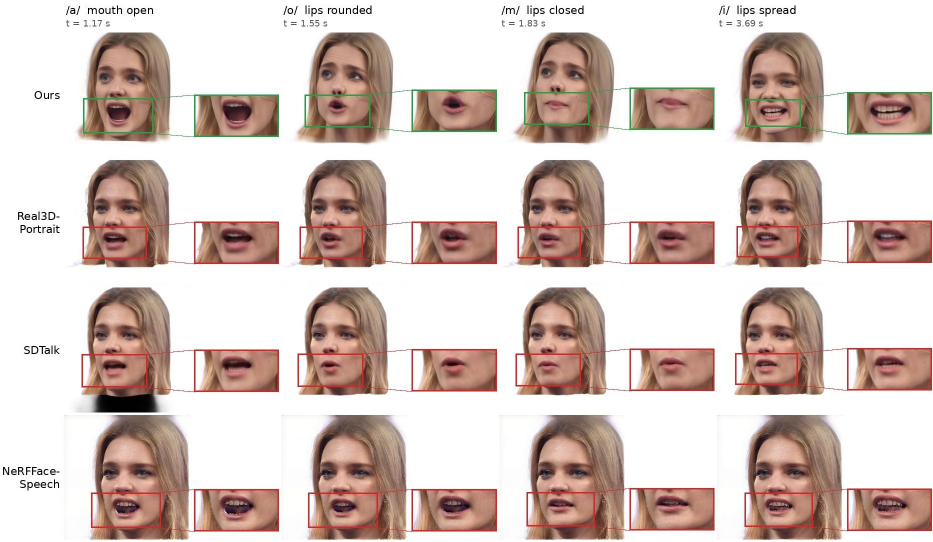}
\caption{Photographic portrait, sung Japanese.}
\label{fig:itw_a9}
\end{figure}

\begin{figure}[htbp]
\centering
\includegraphics[width=0.62\linewidth]{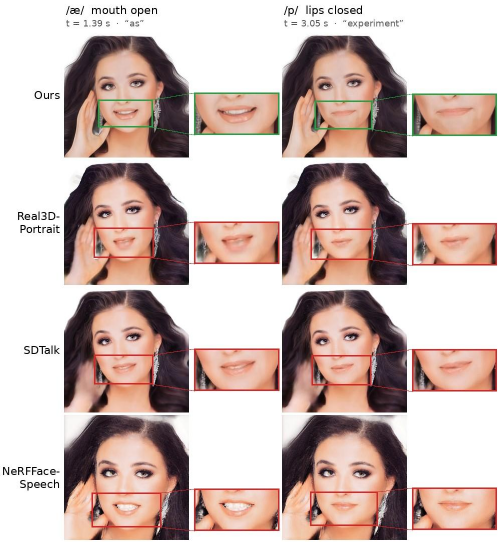}
\caption{Portrait with drop earrings and a raised hand, English speech.}
\label{fig:itw_a10}
\end{figure}

\clearpage
\subsection{Temporal Consistency}
\label{app:itw_temporal}

Each strip follows one clip over a long window, so that identity,
accessories, and background can be checked for drift across time rather
than at a single instant.

\begin{figure}[htbp]
\centering
\includegraphics[width=\linewidth]{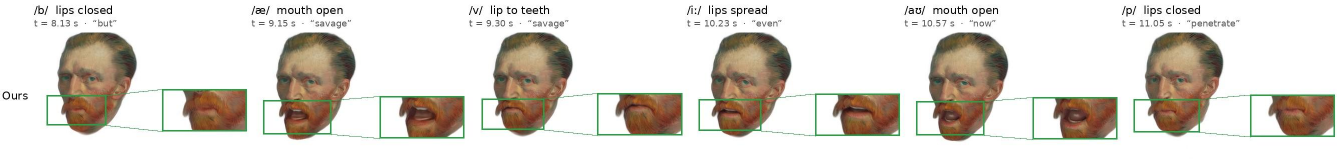}
\\[1pt] {\small (a)}\\[8pt]
\includegraphics[width=\linewidth]{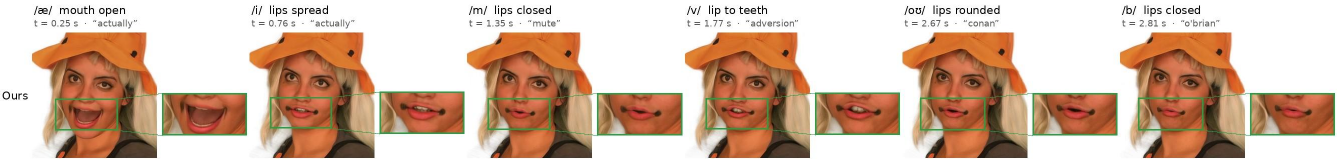}
\\[1pt] {\small (b)}\\[8pt]
\includegraphics[width=\linewidth]{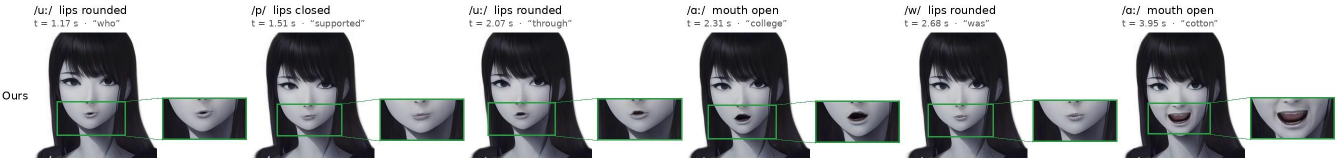}
\\[1pt] {\small (c)}
\caption{Temporal consistency over long windows. (a) Van Gogh
self-portrait, English speech. (b) Wide-brimmed hat and face paint,
English speech. (c) Anime portrait, English speech.}
\label{fig:itw_temporal}
\end{figure}

\subsection{Long-Video Inference}
\label{app:itw_long}

Both clips below are driven by more than 20 seconds of audio (25~s of singing and 22.6~s of speech), well beyond the window
seen during training, to show that quality and identity hold up over long
inputs.

\begin{figure}[htbp]
\centering
\includegraphics[width=\linewidth]{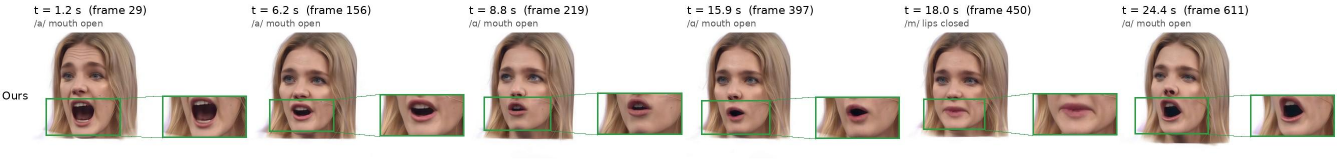}
\\[1pt] {\small (a)}\\[8pt]
\includegraphics[width=\linewidth]{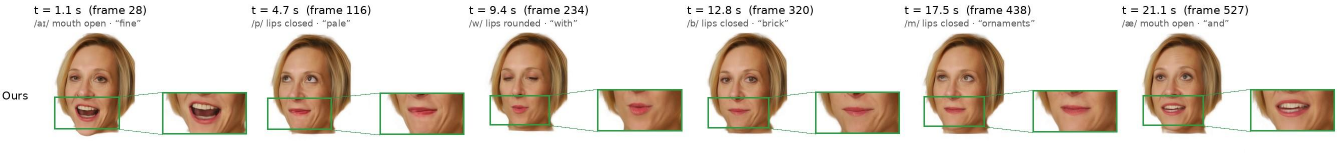}
\\[1pt] {\small (b)}
\caption{More than twenty seconds of driving audio. (a) 25~s of sung Japanese. (b) 22.6~s of English speech.}
\label{fig:itw_long}
\end{figure}

\subsection{Multi-View Consistency}
\label{app:itw_multiview}

Each figure renders the same reconstruction from five azimuths at two time
instants, one where the phoneme requires an open mouth and one where it
requires closed lips. Identity, accessories, and the mouth shape stay
consistent as the camera moves.

\begin{figure}[htbp]
\centering
\includegraphics[width=\linewidth]{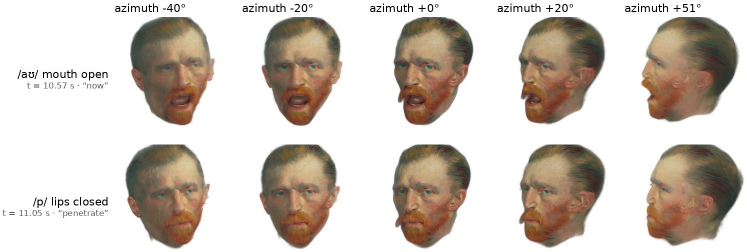}
\caption{Van Gogh self-portrait, English speech, five azimuths.}
\label{fig:itw_d2}
\end{figure}

\begin{figure}[htbp]
\centering
\includegraphics[width=\linewidth]{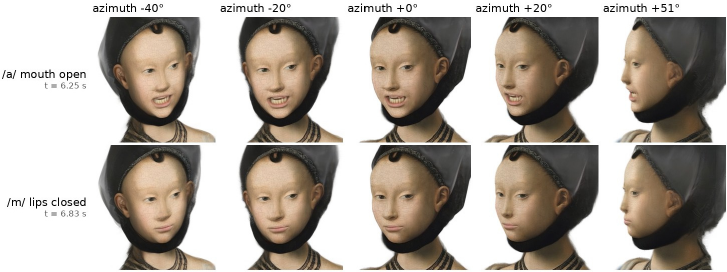}
\caption{Renaissance painting, Swahili speech, five azimuths.}
\label{fig:itw_d3}
\end{figure}

\begin{figure}[htbp]
\centering
\includegraphics[width=\linewidth]{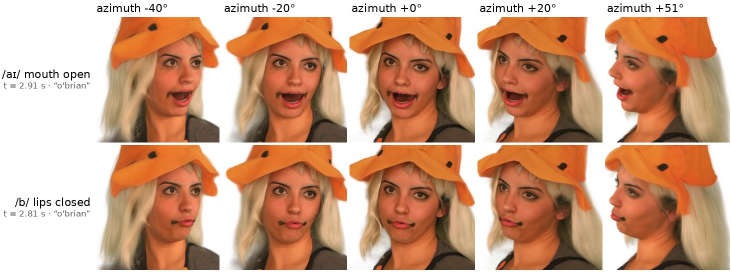}
\caption{Wide-brimmed hat and face paint, English speech, five azimuths.}
\label{fig:itw_d4}
\end{figure}

\begin{figure}[htbp]
\centering
\includegraphics[width=\linewidth]{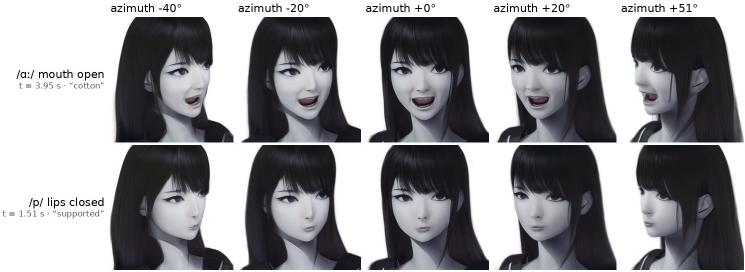}
\caption{Anime portrait, English speech, five azimuths.}
\label{fig:itw_d5}
\end{figure}

\clearpage
\subsection{Sources of the In-the-Wild Pairs}
\label{app:itw_sources}

\paragraph{Scope.}
The in-the-wild results in this appendix and in the supplementary videos
use sixteen image--audio pairs, listed in Table~\ref{tab:itw_sources}. They
come from a pool of 150 pairs that we assembled for qualitative comparison.
We selected the sixteen pairs by visually inspecting our results, so they
are not a random sample. None of the images or recordings belongs to our
training data (NeRSemble) or to our evaluation sets (RenderMe-360 and
HDTF).

\paragraph{Reference images.}
Eight reference images are photographs from FFHQ, chosen by eye from a
seeded random sample of 300 of its 70,000 images. The other eight come from
Wikimedia Commons: five are photographs, two are public-domain paintings
(Vincent van Gogh's \emph{Self-Portrait} of 1887 and Petrus Christus's
\emph{Portrait of a Young Woman}), and one is a public-domain AI-generated
illustration. We searched Commons by keyword and kept only public-domain,
CC0, CC~BY, and CC~BY-SA files. For the photographs and the Christus
portrait, we also required a portrait-oriented image with exactly one
detected face whose center lies in the upper half of the frame, so that the
upper body is visible. We rejected images that show a second person or
whose matte damages the hair or accessories. Each image is prepared like
our training references. Robust Video Matting removes the background,
except for the van Gogh painting and the AI-generated illustration, which
use rembg with U$^2$-Net. The foreground is composited onto white and
framed to $832\times1216$ from MediaPipe face landmarks. The baselines
receive the same square input as in Appendix~\ref{app:baseline_generation}:
the crop $(0,191,832,1023)$ of the $832\times1216$ reference, stored as an
$832\times832$ JPEG at quality 95.

\paragraph{Driving audio.}
All voices are real recordings from public speech and singing corpora: six
LibriSpeech test-clean utterances of English read speech, two FLEURS
utterances of Swahili and French read speech, five GTSinger songs sung in
English, Italian, German, and Japanese, and three Expresso utterances of
English speech read in enunciated, sad, and happy styles. Each GTSinger
recording concatenates, in order, the plain-technique phrases of one song.
All recordings are resampled to 16~kHz mono and peak-normalized to
$-1$~dBFS, with no time-stretching or voice conversion. Voices were
assigned to images by a random draw with a fixed seed that matches the
voice's gender to the apparent gender of the depicted person. Every face
therefore speaks with another person's voice, and no video shows real
speech by the person depicted.

\paragraph{Clip length.}
Recordings of up to 13.5~s are used in full. Longer recordings are cut near
10--13~s, which gives clips of 9.7--12.6~s. Songs end with their first
phrase as segmented in GTSinger, and speech ends at a sentence pause
located by silence detection. All methods receive the same audio file. Two
long-sequence examples use longer audio: the full 22.6-s recording of clip
014 and the first 25.0~s of clip 090.

\paragraph{Licenses and attribution.}
Every image and recording is either in the public domain or under a
Creative Commons license that permits non-commercial reuse and adaptation.
Table~\ref{tab:itw_sources} credits the author and gives the license of
each image, and it gives the corpus, recording, and license of each audio
clip. Links to the images and corpora follow the table. The FFHQ dataset is
distributed under CC~BY-NC-SA~4.0, and each FFHQ image keeps the license
its author chose on Flickr. Our figures and videos modify these works by
background removal, cropping, resizing, and animation. Figures adapted from
CC~BY-SA images are shared under CC~BY-SA~4.0. We use all of these assets
for non-commercial research only.

\begin{table}[t]
\centering\scriptsize
\setlength{\tabcolsep}{3pt}
\caption{Sources and licenses of the in-the-wild pairs. Length is the duration of the driving audio in seconds. The second length of clips 014 and 090 belongs to their long-sequence version. Commons stands for Wikimedia Commons.}
\label{tab:itw_sources}
\vspace{3pt}
\begin{tabular}{@{}r p{0.39\linewidth} p{0.43\linewidth} r@{}}
\toprule
Clip & Reference image & Driving audio & Length \\
\midrule
009 & \raggedright FFHQ \#47591, photograph by Danny Bollinger; CC~BY~2.0 & \raggedright LibriSpeech \texttt{1089-134686-0002}, English; CC~BY~4.0 & 6.6 \\
011 & \raggedright FFHQ \#53017, photograph by Maverick L.; CC~BY~2.0 & \raggedright GTSinger \texttt{IT-Bass-1}, \emph{Di Provenza il mar, il suol}, Italian; CC~BY-NC-SA~4.0 & 9.7 \\
014 & \raggedright FFHQ \#00555, photograph by Joseph Liao; CC~BY-NC~2.0 & \raggedright LibriSpeech \texttt{2094-142345-0000}, English; CC~BY~4.0 & 11.0 / 22.6 \\
020 & \raggedright FFHQ \#58471, photograph by Angelo \& Mantia Photography; CC~BY~2.0 & \raggedright LibriSpeech \texttt{3729-6852-0001}, English; CC~BY~4.0 & 12.2 \\
022 & \raggedright FFHQ \#42543, photograph by April Killingsworth; CC~BY~2.0 & \raggedright GTSinger \texttt{EN-Alto-1}, \emph{someday or oneday}, English; CC~BY-NC-SA~4.0 & 10.6 \\
025 & \raggedright AI-generated illustration (2022) by Benlisquare, Commons; public domain & \raggedright LibriSpeech \texttt{1995-1826-0002}, English; CC~BY~4.0 & 4.6 \\
081 & \raggedright FFHQ \#57539, photograph by Waikiki Natatorium; CC~BY~2.0 & \raggedright GTSinger \texttt{DE-Soprano-1}, \emph{Ich Sage Ja}, German; CC~BY-NC-SA~4.0 & 9.7 \\
085 & \raggedright FFHQ \#62922, photograph by Miha TAMURA; CC~BY-NC~2.0 & \raggedright Expresso \texttt{ex04\_enunciated\_00008}, English, enunciated style; CC~BY-NC~4.0 & 12.1 \\
089 & \raggedright V.~van Gogh, \emph{Self-Portrait} (1887), Commons; public domain & \raggedright LibriSpeech \texttt{1320-122612-0000}, English; CC~BY~4.0 & 13.5 \\
090 & \raggedright FFHQ \#16581, photograph by Keven Menard Photography; CC~BY-NC~2.0 & \raggedright GTSinger \texttt{JA-Soprano-1}, \emph{Hi ga Ochiru Made}, Japanese; CC~BY-NC-SA~4.0 & 12.6 / 25.0 \\
102 & \raggedright P.~Christus, \emph{Portrait of a Young Woman} (c.~1470), Commons; public domain & \raggedright FLEURS \texttt{sw\_ke} test \#1802, Swahili; CC~BY~4.0 & 7.8 \\
106 & \raggedright Photograph by Newpageantfan2020, Commons; CC~BY-SA~4.0 & \raggedright Expresso \texttt{ex02\_sad\_00035}, English, sad style; CC~BY-NC~4.0 & 4.0 \\
110 & \raggedright Photograph by Volha Sh, Commons; CC~BY-SA~4.0 & \raggedright LibriSpeech \texttt{7127-75946-0000}, English; CC~BY~4.0 & 11.4 \\
133 & \raggedright Photograph by Viktoria Borodinova, Commons; CC~BY-SA~4.0 & \raggedright GTSinger \texttt{EN-Alto-2\_2}, \emph{Safe and Sound}, English; CC~BY-NC-SA~4.0 & 12.6 \\
135 & \raggedright Photograph by Kevin Dooley, Commons; CC~BY~2.0 & \raggedright Expresso \texttt{ex04\_happy\_00369}, English, happy style; CC~BY-NC~4.0 & 3.5 \\
138 & \raggedright Photograph by Arne Hodali\v{c}, Commons; CC~BY-SA~4.0 & \raggedright FLEURS \texttt{fr\_fr} test \#1855, French; CC~BY~4.0 & 11.6 \\
\bottomrule
\end{tabular}

\smallskip
\begin{minipage}{\linewidth}\scriptsize\raggedright
Image links: 009~\url{https://www.flickr.com/photos/dannybollinger/8474775720/}; 011~\url{https://www.flickr.com/photos/laubetomas/20415471643/}; 014~\url{https://www.flickr.com/photos/choweee/3920883192/}; 020~\url{https://www.flickr.com/photos/twogeeksandacamera/5277697297/}; 022~\url{https://www.flickr.com/photos/aprillynn77/2186765664/}; 025~\url{https://commons.wikimedia.org/wiki/File:Algorithmically-generated_black_and_white_portrait_art_of_a_young_Japanese_woman_in_the_snow.png}; 081~\url{https://www.flickr.com/photos/natatorium/27339381005/}; 085~\url{https://www.flickr.com/photos/42631759@N05/25897604984/}; 089~\url{https://commons.wikimedia.org/wiki/File:Vincent_van_Gogh_-_Self-Portrait_-_Google_Art_Project_(454045).jpg}; 090~\url{https://www.flickr.com/photos/kevenmenard/5860298556/}; 102~\url{https://commons.wikimedia.org/wiki/File:Petrus_Christus_-_Portrait_of_a_Young_Woman_-_Google_Art_Project.jpg}; 106~\url{https://commons.wikimedia.org/wiki/File:Amelia_Casewell_Headshot.jpg}; 110~\url{https://commons.wikimedia.org/wiki/File:Scientist_(chemist)_Siarhei_Besarab_Studio_portrait_2026_(04).jpg}; 133~\url{https://commons.wikimedia.org/wiki/File:Woman_style_portrait_fashion.jpg}; 135~\url{https://commons.wikimedia.org/wiki/File:Cosplay_2010_portraits_-_Soul_Eater_Eruka_Frog_Cosplayer.jpg}; 138~\url{https://commons.wikimedia.org/wiki/File:Arne_Hodali%C4%8D_in_his_studio_2022.jpg}.

Corpus links: FFHQ~\url{https://github.com/NVlabs/ffhq-dataset}; LibriSpeech~\url{https://www.openslr.org/12}; FLEURS~\url{https://huggingface.co/datasets/google/fleurs}; GTSinger~\url{https://huggingface.co/datasets/GTSinger/GTSinger}; Expresso~\url{https://speechbot.github.io/expresso/}.
\end{minipage}
\end{table}
\FloatBarrier

%% file: SEC/A_metrics.tex
\section{Evaluation Protocols and Metric Implementation}
\label{app:metric_psnr_ssim}

This appendix specifies the evaluation underlying
Tables~\ref{tab:renderme_end_to_end}, \ref{tab:hdtf_end_to_end}, and
\ref{tab:quantitative_ablations}: frontal PSNR, SSIM, LPIPS, and FID;
seven-view CSIM-ref; paired novel-view reconstruction; and the two
quantitative ablations. The datasets, frame ranges, and PSNR masks differ
between the external comparisons and the ablations, as detailed below.

\subsection{Evaluation Cohorts}
\label{app:metric_cohorts}

RenderMe-360 uses one clip from each of two sessions, \texttt{s1\_all} and
\texttt{s2\_all}, for each of 21 identities: 0026, 0041, 0048, 0094, 0099,
0100, 0116, 0156, 0168, 0175, 0189, 0195, 0232, 0250, 0253, 0259, 0262,
0278, 0290, 0295, and 0297. Every method receives the same underlying
reference portrait and 3.24-second, 16-kHz mono speech segment for a clip,
with method-specific input preprocessing. Generated videos contain 81 frames
at 25 fps. Both sessions use the source interval from 5.00 to 8.24 seconds.
At the source rate of 30 fps, output frame $t$ uses raw frame
$150+\operatorname{round}(30t/25)$ for $t=0,\ldots,80$.
The reference portrait comes from source-camera frame 150, the first frame
of this selected window, not the beginning of the recording.

HDTF uses one portrait--audio pair for each of 21 subjects: AnnMarieBuerkle,
AnnWagner, AustinScott0, BillCassidy, BobCorker, BobbySchilling,
CandiceMiller, CarlyFiorina0, CathyMcMorrisRodgers0, ChuckGrassley,
CoryGardner, DanNewhouse, DanSullivan, DaveCamp, DavidVitter, DeanHeller,
DebFischer0, DianeBlack0, DuncanHunter, ErikPaulsen, and FredUpton.
The same 21-subject subset is used for every method and every reported HDTF metric.
Table~\ref{tab:clip_window_selection} summarizes the temporal windows used
for training and external evaluation. Every window contains 3.24 seconds
of 16-kHz mono audio (51,840 samples) and 81 video frames at 25 fps.
The nominal frame times relative to the window start are
$0,0.04,\ldots,3.20$ seconds; the audio spans the full 3.24-second window.

\begin{table}[!htbp]
\centering
\small
\caption{Selection of the 81-frame windows. Start times refer to the source
recording, not to the generated video.}
\label{tab:clip_window_selection}
\begin{tabular}{@{}p{0.24\linewidth}p{0.71\linewidth}@{}}
\toprule
Dataset and use & Selected source interval \\
\midrule
NeRSemble (training) & 0.00--3.24 seconds of each retained SEN recording. \\
RenderMe-360 (evaluation) & 5.00--8.24 seconds of each session, shared by all 42 clips. \\
HDTF (evaluation) & A subject-specific 3.24-second window within an official
target-speaker interval, starting at a speech segment when possible;
for example, 13.74--16.98 seconds for AnnMarieBuerkle. \\
\bottomrule
\end{tabular}
\end{table}

For HDTF, we first select a reference-verified, single-speaker window within
the official target-speaker intervals, favoring acoustically homogeneous
segments without shot changes. Within this window, we take the beginning
of the first pause-delimited speech segment lasting at least 3.24 seconds.
Speech segments are detected from 20-ms audio-energy measurements, with
pauses of at least 0.35 seconds separating them. If no segment is long
enough, we start from the longest available segment and move the start
earlier if needed to keep the full 3.24 seconds inside the selected window.
This fallback can cross a pause and is used for two of the 21 clips.
The reference portrait is verified against the target speaker but need not
come from the evaluation window's first frame. Window selection is based
on source material, not generated videos, and the chosen reference and
audio are shared by all methods.

To pair HDTF predictions with GT, we sample the source timeline at 25 fps
from the chosen audio start: output frame $t$ corresponds to $t/25$ seconds
after that start. We use the decoded source video frame with the nearest
timestamp, choosing the earlier one on ties. Audio start times are relative
to the first decoded audio sample, accounting for any offset between the
audio and video streams rather than assuming that the container timestamp
is their common origin. We do not shift or retime a prediction to improve
its agreement with GT.

\subsection{Shared Frontal Evaluation Protocol}
\label{app:metric_frontal_shared}

For both datasets, we exclude frame 0 because it is the
reference-conditioned anchor rather than an ordinary audio-driven frame;
including this easier frame can inflate the clip mean. We apply the same
exclusion to every method. For each remaining frame
$t\in\{1,\ldots,80\}$, we compare the predicted frontal output with the
ground-truth observation at the corresponding time: camera 25 on
RenderMe-360 and the source video on HDTF. Each method therefore contributes
$42\times80=3{,}360$ paired frontal comparisons on RenderMe-360 and
$21\times80=1{,}680$ on HDTF.

All methods use the same frame indices, ground truth, scoring resolution, and
metric implementation. Their native outputs require different deterministic
mappings to the common $512\times512$ RGB canvas:
\begin{center}
\small
\begin{tabular}{lll}
\hline
Method & Native output & Mapping to the $512\times512$ scoring canvas \\
\hline
Ours & Stage-II frontal render, $832\times1216$ &
crop to $832\times832$, then Lanczos resize \\
Real3D-Portrait & $512\times512$ & use directly \\
SDTalk & $512\times512$ & prediction-only background normalization \\
NeRFFaceSpeech & aligned $1024\times1024$ & Lanczos resize only \\
\hline
\end{tabular}
\end{center}

For our method, the evaluated video is the frontal rendering of the final
Stage-II reconstructed avatar, not the intermediate Stage-I frontal video.
We crop the portrait with
$(x_0,y_0,x_1,y_1)=(0,191,832,1023)$, obtaining an
$832\times832$ square, and resize it to $512\times512$ with Pillow Lanczos.
Real3D-Portrait and SDTalk already produce $512\times512$ videos.
NeRFFaceSpeech uses its released face-alignment preprocessing with the
largest detected source face selected consistently for all inputs. Its
decoded $1024\times1024$ output is resized directly to $512\times512$;
we do not invert its face alignment or fit the output to the GT framing.
Thus a common scoring resolution does not imply pixel-aligned facial crops
across methods. Frontal scores use decoded native videos, not annotated
figures, side-by-side previews, or intermediate Stage-I outputs.

For RenderMe-360 GT, we transform raw camera-25 RGB and the official soft
foreground mask independently using the same clip-fixed portrait window,
white RGB padding and zero alpha padding, Lanczos resizing to
$832\times1216$, the square crop above, and Lanczos resizing to
$512\times512$. Only then do we composite GT once over white. For uint8
RGB $C$ and alpha $a$, the stored value is
$\lfloor(Ca+255(255-a)+127)/255\rfloor$, with alpha broadcast over channels.
HDTF follows the same fixed-window and square-crop procedure but uses an
estimated GT alpha from Robust Video Matting (ResNet-50), not an official
dataset mask. Its recurrent state is reset per clip and advanced over all
81 selected source frames, with downsampling ratio
$\min(512/\max(W,H),1)$ on the raw source image. We retain raw GT RGB,
quantize the predicted alpha to uint8, transform RGB and alpha separately,
and composite once at $512\times512$. Framing is fixed from the anchor
and does not track the generated face.

SDTalk's native black background is normalized using a fixed
\texttt{BiRefNet-matting} model applied only to its prediction. The RGB input
is bilinearly resized to $1024\times1024$ and ImageNet-normalized; the final
sigmoid matte is converted to uint8 and bicubically resized to
$512\times512$. Given native RGB $P$ and estimated alpha $\hat a$, the
scored prediction is $\operatorname{clip}(P+255-\hat a,0,255)$, with the
addition performed before conversion to uint8. This is an approximate
black-to-white replacement, not recovered renderer alpha. The model and
rule are shared across clips and views, without GT masks, thresholding,
morphological corrections, or per-example tuning. Errors in this external
matte remain in the scored images. It is not applied to CSIM-ref videos.
For all methods, predictions are not masked with GT alpha or spatially
aligned to each GT frame; PSNR, SSIM, LPIPS, and FID use the whole resulting
canvas. This differs from the earlier jointly GT-masked scoring variant.

For PSNR, SSIM, and LPIPS, we first average the 80 frame scores within a clip, average
the two sessions belonging to each identity, and then average the 21
identities with equal weight on RenderMe-360. On HDTF, each subject has one
clip, so the 21 clip means are averaged equally. FID uses the set-level
procedure described below.

\subsection{Frontal PSNR}
\label{app:metric_psnr_v7}

For a paired prediction $P_t$ and ground truth $G_t$, represented as uint8
RGB arrays over all $512\times512\times3$ elements, we compute
\[
  \mathrm{MSE}_t
  = \operatorname{mean}\!\left[(P_t-G_t)^2\right],
  \qquad
  \mathrm{PSNR}_t
  = 10\log_{10}\frac{255^2}{\mathrm{MSE}_t}.
\]
Squared errors are computed after conversion to floating point to avoid
uint8 subtraction overflow, with a peak RGB value of 255.
It computes PSNR independently for every frame and then averages the resulting
dB values; it does not average MSE over the entire clip before taking the
logarithm.

\subsection{Frontal SSIM}
\label{app:metric_ssim_v7}

SSIM uses exactly the same prediction--GT pairs, common canvas,
frames 1--80, and hierarchical averaging as F-PSNR. Only the image comparison
function changes. For two local RGB windows $x$ and $y$, it computes
\[
  \mathrm{SSIM}(x,y)=
  \frac{(2\mu_x\mu_y+C_1)(2\sigma_{xy}+C_2)}
       {(\mu_x^2+\mu_y^2+C_1)(\sigma_x^2+\sigma_y^2+C_2)},
  \quad
  C_1=(0.01\cdot255)^2,\quad C_2=(0.03\cdot255)^2.
\]
We call scikit-image's \texttt{structural\_similarity} with
\texttt{channel\_axis=2} and \texttt{data\_range=255}.
The audited implementation uses its default $7\times7$ uniform window and
sample covariance, and aggregates spatial locations and RGB channels into
one scalar per frame.

\subsection{Frontal LPIPS}
\label{app:metric_lpips_v8}

LPIPS uses the same frame pairs, common $512\times512$ canvases,
frames 1--80, and hierarchical averaging as F-PSNR and F-SSIM.
Each RGB uint8 image $X$ is converted to
\[
  X_{\mathrm{LPIPS}}=X/127.5-1,
\]
giving values in $[-1,1]$, and is rearranged from HWC to NCHW.
The evaluator calls \texttt{lpips.LPIPS(net=alex, version=0.1)}
with the fixed AlexNet weights in evaluation mode and without gradients.
It outputs one distance for each paired frame. These frame distances are
aggregated using the dataset-specific rule in
Appendix~\ref{app:metric_frontal_shared}.

\subsection{Frontal FID}
\label{app:metric_fid_v8}

FID uses the same common-canvas frames 1--80, but treats them as two
image sets rather than paired comparisons. For each method, prediction and
GT each contain $42\times80=3{,}360$ images on RenderMe-360 and
$21\times80=1{,}680$ on HDTF. The evaluator uses clean-fid in clean mode
to extract a 2,048-dimensional Inception feature for every image.
If the two feature sets have means $\mu_p,\mu_g$ and sample covariances
$\Sigma_p,\Sigma_g$, it computes
\[
  \mathrm{FID}=
  \|\mu_p-\mu_g\|_2^2+
  \operatorname{Tr}\!\left(
  \Sigma_p+\Sigma_g-2(\Sigma_p\Sigma_g)^{1/2}\right).
\]
This produces one pooled FID value for each method; there is no per-frame
or per-clip FID that is subsequently averaged.

GT and prediction features for each FID calculation use the same extractor
weights, preprocessing, and execution backend. On HDTF, the pooled mean and
covariance are computed from the 1,680 feature vectors belonging to the
21 evaluated subjects, separately for GT and each method; we do not average
per-subject FIDs. The tables report point estimates without FID confidence
intervals.

\subsection{Reference-Image Identity Similarity (CSIM-ref)}
\label{app:metric_csim_ref_v8}

CSIM-ref compares each generated face with the reference image actually
consumed by that method, rather than with a time-matched GT frame; on HDTF,
all methods are compared with the same $832\times1216$ reference portrait, of
which the baselines' square input is a crop. It uses
native decoded frames before the common-canvas conversion and evaluates
frames 1--80. RGB images are converted to BGR and passed to InsightFace
\texttt{FaceAnalysis} with the \texttt{buffalo\_l} model, CPU execution,
and detector input size $(320,320)$ instead of the library default
$(640,640)$. For each frame, we compute CSIM using
the largest detected face. The returned
\texttt{normed\_embedding} is explicitly L2-normalized once more.

The detector input size only sets the resolution at which the detector
searches for faces; the detection threshold ($0.5$), the five-landmark
alignment to a $112\times112$ crop, and the recognition network are
unchanged. It matters only for faces that fill most of the frame, such as
NeRFFaceSpeech's $1024\times1024$ face-aligned outputs, especially in side
views: at $(640,640)$ these faces are scored below the threshold and
dropped, whereas at $(320,320)$ the same faces are detected. Because a
smaller input size could in principle admit a wrong detection, we keep
scoring only the largest detected face in each frame. We verified on the
same RenderMe-360 videos (four methods, seven views, 42 clips) that the
change is safe. No face detected at $(640,640)$ is lost at $(320,320)$, and
where both settings detect a face, the selected boxes overlap with an IoU of
at least $0.6$ in every frame ($0.94$ on average), so the same face is
scored. Because the facial landmarks used for alignment shift slightly
between the two input sizes, individual frame scores fluctuate slightly, but
the fluctuations are equally likely to go up or down and cancel out: the
final scores of Ours, Real3D-Portrait, and SDTalk change by at most $0.003$,
and the conclusions are unaffected. The only substantive effect is that
6,508 NeRFFaceSpeech frames missed at $(640,640)$ are now scored, which is desired.

Let $e_t$ and $e_{\mathrm{ref}}$ denote the normalized embeddings of a
predicted frame and its method-specific reference image. The frame score is
\[
  \mathrm{CSIM\mbox{-}ref}_t=e_t^\top e_{\mathrm{ref}}.
\]
Frames without a detected face are excluded from the clip mean, and we report
the detection coverage.

We report a multi-view CSIM-ref. In addition to the frontal video, each
method renders the same audio-driven sequence with the camera rotated by
$\pm20^\circ$, $\pm40^\circ$, and $\pm60^\circ$ of yaw about the head at
unchanged elevation, using each method's own camera convention, giving seven
81-frame talking videos per clip. Valid frame scores are averaged within
each video, and the clip value is the equal-weight mean of the seven
per-view means. On RenderMe-360, clip values are averaged over the two
sessions per identity and then over the 21 identities with equal weight; on
HDTF, they are averaged over the 21 selected subjects.

\subsection{Paired Novel-View PSNR, SSIM, and LPIPS}
\label{app:metric_novel_view_v8}

We evaluate novel-view fidelity at anchor frame 0 using RenderMe-360 cameras
22, 28, 23, 26, and 29. For each of the 42 clips, a method's rendering at each
registered target camera is paired with the simultaneous GT observation from
that camera, giving $42\times5=210$ image pairs per evaluated method. The
comparison includes all four methods: ours, Real3D-Portrait, SDTalk, and
NeRFFaceSpeech. Camera 25 is the source-camera diagnostic and is not one of
these five targets; other rendered camera groups are not pooled into this
table entry.

For every target camera, raw RGB and official soft alpha are independently
cropped and Lanczos-resized using the frozen camera-specific framing, then
GT is composited once over white at $512\times512$. The target intrinsics
are transformed with the same crop/resize. Predictions use the corresponding
registered camera and fixed method adapter, without a GT foreground mask;
SDTalk uses the prediction-only background normalization described above.
NeRFFaceSpeech renders at $1024\times1024$ with the corresponding full
intrinsics and is Lanczos-downsampled to $512\times512$.

Registration uses the physical source camera and source-frame 3D calibration
landmarks to establish scale, orientation, and source framing in each
model's coordinate system, then transfers the resulting transform to the
target cameras. These landmarks are evaluation calibration metadata, not
additional RGB reconstruction inputs. The per-clip registration uses source
observations rather than target-image alignment. Method-level calibration
choices and the fixed adapters are specified in
Appendix~\ref{app:baseline_cameras}; target crop windows remain fixed across
methods. For each clip we retain the same
frame-0 reconstruction or fitted expression state across target views,
including each baseline's released view-dependent rendering operations.

Novel-view PSNR, SSIM, and LPIPS use the same per-image implementations as
their frontal counterparts above. We first average the five target-camera
scores within each clip, then average the two sessions per identity, and
finally average the 21 identities with equal weight. The table reports these
point estimates. No novel-view FID is included, and HDTF has no paired
novel-view PSNR, SSIM, or LPIPS because it lacks synchronized target cameras.

\input{SEC/A_ablation_evaluation}

\subsection{Mouth-Opening Contrast}
\label{app:mouth_opening_contrast}

We measure mouth opening as the central inner-lip distance divided by the
mouth-corner distance, using the image coordinates of InsightFace
\texttt{buffalo\_l} 68-point landmarks on $512\times512$ frontal frames
and selecting the largest detected face. We greedily decode phonemes from
the driving audio with \texttt{wav2vec2-lv-60-espeak-cv-ft}, retain tokens
with peak posterior at least 0.5, and sample the nearest 25-fps frame to
each token-run midpoint, shifted 40\,ms earlier for \mbox{/b/} and
\mbox{/p/}. The contrast is the mean normalized opening at
\mbox{/a/}-type open vowels minus that at \mbox{/m/}, \mbox{/b/}, and
\mbox{/p/}, pooling valid tokens across the 16 clips rather than averaging
clip means. For each method, we search one shared integer offset across
all clips within $[-6,6]$ frames and report the maximum contrast; this is
not a separate alignment for each clip. Missing detections and out-of-range
frames are excluded. At the selected offsets, all methods have 82 open-vowel
and 81 bilabial observations, with none missing. This is an auxiliary
diagnostic of audio-conditioned opening--closing contrast.

%% file: SEC/A_ablation_evaluation.tex
\subsection{Quantitative Ablation Protocols}
\label{app:quantitative_ablation_evaluation}

Both parts of Table~\ref{tab:quantitative_ablations} use eight NeRSemble
identities disjoint from training: 145, 222, 279, 302, 369, 413,
476, and 490, with one clip per identity. The respective sequence/clip pairs
are SEN-01/0021, SEN-02/0000, SEN-01/0015, SEN-01/0045, SEN-01/0000,
SEN-01/0030, SEN-01/0060, and SEN-01/0003. Predictions are paired
with GT from the same identity, time index, and target camera. These internal
ablations use $416\times608$ RGB images rather than the external benchmarks'
$512\times512$ canvas.

\paragraph{Fifth frontal view.}
Table~\ref{tab:stage1_four_five_views}(a) fixes the final five-view Stage-I
generator, its four synthesized grid views, and the released, frozen FaceLift
reconstructor. The five-view condition adds the frontal output of that same
generator; it does not add a GT image or use a separately trained generator.
We score the resulting reconstructions from frontal camera
\texttt{222200037} at frames 0--80, giving $8\times81=648$ image pairs per
condition. The archived prediction videos were encoded from $416\times608$
renders, and GT was Lanczos-downsampled from $832\times1216$ to the same size;
both use H.264, CRF 10, \texttt{yuv420p}, and 25 fps. We decode the archives
to RGB and pair frames by index, using OpenCV Lanczos4 to resize a prediction
to the GT dimensions only if needed. No temporal or facial registration is
performed. The PSNR column is \emph{foreground-only}: its MSE uses RGB errors
at GT pixels satisfying $R+G+B<735$ in the decoded white-background image.
SSIM and LPIPS use the entire image, not this foreground mask.

\paragraph{View-aware corruption.}
Table~\ref{tab:stage2_checkpoint_comparison}(b) compares released FaceLift,
the checkpoint after 3,240 random-camera/background-adaptation updates, and
its continuation for 3,240 updates with view-aware corruption, all from the
16-camera training lineage. All three receive the same
five Stage-I-generated inputs at every time index: the four $416\times608$
grid cells for cameras \texttt{222200042}, \texttt{221501007},
\texttt{222200044}, and \texttt{222200045}, and the generated frontal view
for \texttt{222200037}, fitted to the same canvas with aspect-preserving
Lanczos resizing and white padding. No synthetic corruption is added at
evaluation time. Each checkpoint reconstructs one Gaussian cloud per frame,
then renders that cloud at all 16 calibrated cameras: \texttt{220700191},
\texttt{221501007}, and \texttt{222200036}--\texttt{222200049}. These are
training-pool cameras, not the four held-out cameras of the 12-camera variant
in Appendix~\ref{app:stage2_camera_pools}. We evaluate frames 1--80,
excluding the anchor, for $8\times80\times16=10{,}240$ image pairs per row.
All rows use the same renderer and model configuration, without post-hoc
Gaussian filtering. Predictions are rendered on white, clamped to $[0,1]$,
and rounded to uint8 RGB. They are compared directly with the existing
matted, white-background GT at $416\times608$, with no additional crop,
registration, or GT mask applied to predictions. All three metrics use
the full image, including any floating artifacts outside the GT silhouette.

\paragraph{Metric implementation and aggregation.}
We compute per-image PSNR from float64 RGB squared errors, SSIM with
\texttt{channel\_axis=2}, \texttt{data\_range=255}, and the default uniform
$7\times7$ window, and AlexNet LPIPS v0.1 on RGB values mapped to $[-1,1]$.
For each identity, we average its 81 frame scores in part (a), or its
$80\times16$ frame--camera scores in part (b), then average the eight
identity means with equal weight. PSNR is averaged in dB after each image's
MSE is converted to PSNR, not computed from one pooled MSE. Aggregation uses
unrounded values; the table shows two decimal places for PSNR and four for
SSIM and LPIPS.

\subsection{Generalization to Held-Out Cameras}
\label{app:heldout_camera_generalization}

We additionally evaluate whether the Stage-II improvements extend beyond the
training viewpoints. The 12-camera variant excludes cameras
\texttt{222200046}, \texttt{222200040}, \texttt{222200049}, and
\texttt{222200039} from both source-view sampling and real-target supervision
throughout training. We use the same eight training-disjoint
identities and fixed five Stage-I-generated input views as in
Appendix~\ref{app:quantitative_ablation_evaluation}. At each time index, each
checkpoint reconstructs one Gaussian cloud, which is rendered at all four
held-out cameras and compared with their simultaneous GT observations.
The full-camera model used for the external comparisons is not used in this
experiment, since these cameras are not held out for that model.

Table~\ref{tab:heldout_camera_generalization} reports the anchor frame
separately from the audio-driven frames: frame 0 gives $8\times4=32$ image
pairs per checkpoint, and frames 1--80 give $8\times80\times4=2{,}560$.
All scores use the full $416\times608$ RGB image on white, the same renderer
and calibrated cameras, and no post-hoc Gaussian filtering or GT masking of
predictions. PSNR, SSIM, and LPIPS follow the implementation above; we first
average over the four cameras and the selected frames within each identity,
then average the eight identity means equally. The frontal camera is not
included in this table. Scores on frames 1--80 also reflect differences
between the generated and recorded facial performance, so they are kept
separate from the anchor-frame comparison.

\begin{table}[!htbp]
\centering
\small
\setlength{\tabcolsep}{4pt}
\caption{Generalization to four held-out NeRSemble cameras using the
12-camera training variant. All checkpoints receive the same generated
five-view inputs. Frame 0 and frames 1--80 are evaluated separately.
Bold indicates the best result among the first three rows; the full
12-camera model is listed separately for reference.}
\label{tab:heldout_camera_generalization}
\begin{tabular}{lcccccc}
\toprule
& \multicolumn{3}{c}{Frame 0} & \multicolumn{3}{c}{Frames 1--80} \\
\cmidrule(lr){2-4}\cmidrule(lr){5-7}
Model & PSNR$\uparrow$ & SSIM$\uparrow$ & LPIPS$\downarrow$
      & PSNR$\uparrow$ & SSIM$\uparrow$ & LPIPS$\downarrow$ \\
\midrule
Released FaceLift & 17.92 & 0.7473 & 0.2627 & 17.25 & 0.7291 & 0.2831 \\
Fine-tuned w/o corruption & 17.73 & 0.7515 & 0.2466 & 17.09 & 0.7328 & 0.2672 \\
Fine-tuned w/ corruption & \textbf{18.00} & \textbf{0.7545} & \textbf{0.2412}
  & \textbf{17.28} & \textbf{0.7339} & \textbf{0.2635} \\
\midrule
Full 12-camera model & 18.03 & 0.7543 & 0.2442
  & 17.31 & 0.7336 & 0.2666 \\
\bottomrule
\end{tabular}
\end{table}

The full 12-camera model improves all three mean metrics over released
FaceLift in both temporal splits, supporting generalization beyond the
training cameras. The corruption-trained checkpoint also improves all three
metrics over the preceding fine-tuned checkpoint. The full model has the
highest PSNR, while the
corruption-trained intermediate checkpoint has slightly better SSIM and
LPIPS.

%% file: SEC/A_stage1_implementation.tex
\clearpage
\numberwithin{equation}{section}
\section{Stage-I Implementation Details}
\label{app:stage1_implementation}

This appendix expands the input construction, token layout, and
audio conditioning of Section~\ref{sec:method_stage1}. Stage I builds on
StableAvatar's audio-conditioned Wan2.1-1.3B diffusion
backbone~\citep{wan2025,tu2025stableavatar}, with hidden width $D=1536$ and
30 transformer blocks. We use $t$ for a latent video position
and $\sigma$ for flow time; the reference context occupies an extra position
$t=0$ and is not part of the generated video. This model-context index is
separate from the output-video frame indices used in the evaluation protocol.

\subsection{Clean Reference-Frame Conditioning}
\label{app:stage1_reference}

\begin{figure}[!htbp]
\centering
\includegraphics[width=\linewidth]{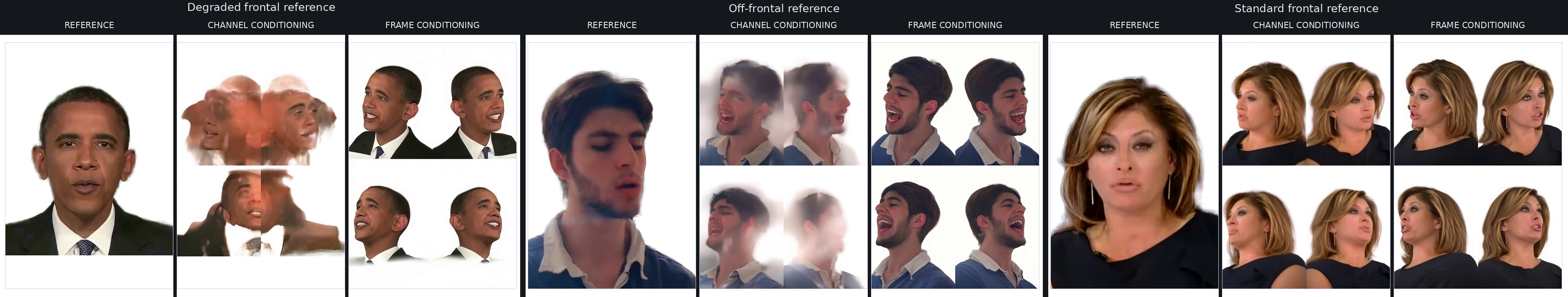}
\caption{Reference-conditioning robustness for degraded frontal, off-frontal,
and standard frontal references. Both strategies produce plausible outputs for
the standard input, with frame-wise conditioning retaining sharper details;
its robustness advantage becomes more pronounced for the two challenging
references.}
\label{fig:app_stage1_identity_robustness}
\end{figure}

StableAvatar injects VAE-based reference information through auxiliary
conditioning channels aligned with the noisy video latent. We instead encode
the reference portrait once with the frozen VAE at canvas resolution as an
unnoised, context-only latent $z_{\mathrm{ref}}$ and prepend it to the noisy
canvas sequence. StableAvatar's CLIP image-conditioning branch is retained.
Let $z_{c,\sigma}$ and $z_{f,\sigma}$ denote the noised VAE latents of the canvas
and frontal video, respectively. We form the temporally aligned streams
\begin{equation}
  \tilde{z}_c=[z_{\mathrm{ref}};z_{c,\sigma}]_T,
  \qquad
  \tilde{z}_f=[\mathbf 0;z_{f,\sigma}]_T,
  \label{eq:app_stage1_reference}
\end{equation}
where $[\cdot;\cdot]_T$ denotes temporal concatenation. The first entry is a
shared context index: $z_{\mathrm{ref}}$ is the reference anchor in the canvas
stream, and $\mathbf 0$ is a zero latent placeholder with the frontal
stream's spatial dimensions. This placeholder aligns the two streams at the
context index. Once their tokens are concatenated, both streams access the
reference through global self-attention. We exclude predictions at the
context index from the training loss and discard them before each inference
scheduler update, keeping the reference latent and frontal placeholder fixed
throughout denoising. The VAE decodes only the target video latents, so the
reference serves as a shared identity anchor rather than an output frame.
Figure~\ref{fig:app_stage1_identity_robustness} compares frame-wise and
channel-wise reference conditioning. For standard frontal portraits, both
conditioning strategies produce plausible multi-view outputs, although
frame-wise conditioning retains sharper facial details. Its advantage becomes
more pronounced for low-resolution or degraded portraits and off-frontal
references, where channel-wise conditioning may produce faded or mixed facial
structures while frame-wise conditioning preserves a coherent identity across
views.

\subsection{Token Layout, Positional Coordinates, and Decoding}
\label{app:stage1_layout}

The canvas and frontal videos have resolutions $832\times1216$ and
$416\times608$ pixels, respectively (width $\times$ height). The frozen VAE
encodes each into 16-channel latents, with $T_\ell=21$ latent frames for an
81-frame video. We append 20 zero auxiliary channels to match the pretrained
36-channel input and project non-overlapping $1\times2\times2$ latent patches
to $D=1536$-dimensional tokens. Adding the context position gives 22
model-time groups. Table~\ref{tab:app_stage1_shapes} summarizes the resulting
shapes; latent dimensions are batch, channels, time, height, width, and
unflattened token dimensions are batch, time, tokens, hidden width.

\begin{table}[t]
  \centering
  \small
  \caption{Stage-I tensor shapes for an 81-frame clip, with batch size $B$.
  Token shapes include one additional context position: a clean reference
  latent for the canvas and a zero latent for the frontal stream, each with
  the corresponding stream's spatial dimensions.}
  \label{tab:app_stage1_shapes}
  \begin{tabular}{ll}
    \toprule
    Tensor & Shape \\
    \midrule
    Canvas video latent & $B\times16\times21\times152\times104$ \\
    Frontal video latent & $B\times16\times21\times76\times52$ \\
    Canvas tokens, with context & $B\times22\times3952\times1536$ \\
    Frontal tokens, with context & $B\times22\times988\times1536$ \\
    Flattened transformer input & $B\times108680\times1536$ \\
    \bottomrule
  \end{tabular}
\end{table}

\paragraph{Frame-wise ordering.}
We reshape each stream's tokens into temporal groups, concatenate the canvas
and frontal blocks along the within-group token dimension, and then flatten
time and token dimensions into one sequence. The order is
$[P_c^0\Vert P_f^0\Vert P_c^1\Vert P_f^1\Vert\cdots]$.
Thus each consecutive block of 4940 tokens belongs to one latent time. This
is block-wise ordering by time, not alternation of individual canvas and
frontal tokens. Appending the entire frontal sequence after the entire canvas
sequence would not preserve the inherited equal-size, per-time audio
partitions without changing that routing logic.

\paragraph{RoPE positions.}
The canvas token map has height 76 and width 52; the frontal token map has
height 38 and width 26. At each model time $t$, canvas tokens retain positions
$(t,h,w)$ with $0\leq h<76$ and $0\leq w<52$. Frontal tokens use
$(t,h,w+52)$ with $0\leq h<38$ and $0\leq w<26$. Consequently, tokens from
the same time share a temporal coordinate while occupying distinct spatial
positions. We construct these coordinates in exactly the same order as the
token sequence and apply the backbone's temporal, height, and width RoPE
frequency bands. No additional view embedding is active in the final model.
These are positional coordinates for attention, not camera extrinsics or an
explicit 3D encoding. Global self-attention can exchange information across
views and times.

\paragraph{Output reconstruction.}
The shared head maps each token to 64 values, corresponding to a
$1\times2\times2$ patch with 16 latent channels. We reshape its output into
22 groups of 4940 tokens, split each group into the first 3952 canvas tokens
and the remaining 988 frontal tokens, and unpatchify each onto its own latent
grid. After discarding the context predictions, the outputs supervise or
update the respective 21-frame latent sequences. The two streams use matching
denoising schedules. The frozen VAE decodes the completed canvas and frontal
latents separately; splitting the decoded canvas into its four cells gives
the four oblique videos alongside the fifth frontal video.

\subsection{Audio, Image, and Text Conditioning}
\label{app:stage1_audio}

\paragraph{Acoustic features and local windows.}
A frozen Wav2Vec2-Base encoder extracts 768-dimensional speech features at a
16 kHz audio sampling rate. The inherited audio adapter first applies a
bias-free linear projection to width 1536 followed by layer normalization.
It then constructs one local feature window per video latent time using the
original RGB-frame timeline and the VAE's temporal compression. Each window
is expanded by four audio-token positions on either side. Indices beyond
the feature sequence are padded with zeros to form a rectangular batch.
For an unpadded 3.24-second waveform corresponding to 81 frames at 25 fps,
the encoder receives 51,840 samples and returns 161 features. The inherited
windowing rule yields 21 windows of 15 positions each, giving a tensor of
shape $B\times21\times15\times1536$. Boundary padding is appended on the
right after slicing the valid features and remains unmasked in attention.

\paragraph{Visual-to-audio refinement.}
Two frozen audio-adapter blocks refine the projected windows. Here audio
tokens are queries, while the corresponding noisy canvas patch tokens are
keys and values. The visual context comes from patch embedding before the
first diffusion-transformer block. We remove the clean reference group and
select only the canvas portion of every target-time group, giving
$B\times21\times3952\times1536$ visual context. Each adapter block uses
eight attention heads of width 192 and a feed-forward layer of width 3072.
The diffusion-timestep embedding modulates its normalized features and
residual branches, and a final modulated head produces the refined audio
context.

\paragraph{Audio-to-visual injection.}
We prepend an all-zero audio context for $t=0$, producing
$B\times22\times15\times1536$ contexts in the standard example. Within each
of the 30 diffusion-transformer blocks, the visual tokens are queries and
the refined audio features are keys and values. Both canvas and frontal
tokens at time $t$ query that time's audio context. Reshaping time into the
batch dimension implements this routing, using 12 heads of width 128.
Each block's audio cross-attention weights are shared across all time groups.
The audio adapter runs once per diffusion-transformer forward pass; its
output is reused across all 30 blocks and recomputed at the next denoising
step.

\paragraph{Other conditioning paths.}
The frozen image encoder supplies portrait features to the inherited image
cross-attention pathway, separately from the VAE reference context in
Appendix~\ref{app:stage1_reference}. The frozen text encoder provides the
prompt context. Within a diffusion-transformer block, the text, image, and
audio attention outputs are summed before the shared cross-attention output
projection. We inherit these conditioning modules from StableAvatar.

\subsection{Training Scope and Loss Details}
\label{app:stage1_training}

\paragraph{Trainable modules.}
We obtain the four-view frame-conditioned warm start by fine-tuning only
transformer self-attention from the released StableAvatar weights
(Table~\ref{tab:stage1_training_schedule}). The five-view model is initialized
from this warm start and additionally trains patch embedding and the output
head, giving approximately 283.7 million trainable parameters.
The VAE, text/image/audio encoders,
audio adapter, text/image/audio cross-attention, and transformer feed-forward
layers remain frozen in both models.

\paragraph{Two-stream flow matching.}
For each stream $s\in\{c,f\}$, training forms
$z_{s,\sigma}=(1-\sigma)z_s+\sigma\epsilon^s$ with target
$u^s=\epsilon^s-z_s$, where $z_s$ is the encoded target video and
$\epsilon^s$ is Gaussian noise. The streams use independently drawn noise
but the same sampled flow time $\sigma$. The clean reference and zero frontal
context are excluded from supervision.

For each training sample, one random draw selects a spatial weighting mode
shared by both streams. The default probabilities are 0.4 for full-frame,
0.1 for face-focused, and 0.5 for lip-focused supervision. Full-frame
weighting uses $1+m_{\mathrm{face}}+m_{\mathrm{lip}}$, while the other two
modes use their corresponding masks. Each stream uses its own masks resized
to its latent resolution with trilinear interpolation. If a lip mask is
absent or empty, its face mask is used instead. The weighted squared error
is averaged over the complete latent tensor, rather than normalized by only
the masked region; the inherited flow-time weighting is also applied.
The two stream losses are summed with frontal coefficient 1.0.

\paragraph{Inherited auxiliary term.}
We retain StableAvatar's auxiliary difference loss on adjacent latent
channels of the canvas prediction. With $\Delta_{\mathrm{ch}}$ denoting
that channel difference
and $v_\theta^c$ the canvas flow prediction after context removal, the actual
combined objective is
\begin{equation}
  \mathcal L_{\mathrm{train}}
  =0.75\bigl(\mathcal L_c+\mathcal L_f\bigr)
   +0.25\operatorname{MSE}
     \!\left(\Delta_{\mathrm{ch}}v_\theta^c,
              \Delta_{\mathrm{ch}}u^c\right).
  \label{eq:app_stage1_full_loss}
\end{equation}
The auxiliary term is an unmasked mean squared error and is applied only to
the canvas stream.

\input{SEC/stage1_sparse_token_architecture}

%% file: SEC/stage1_sparse_token_architecture.tex
\begin{figure*}[t]
  \centering
  \includegraphics[width=\textwidth]{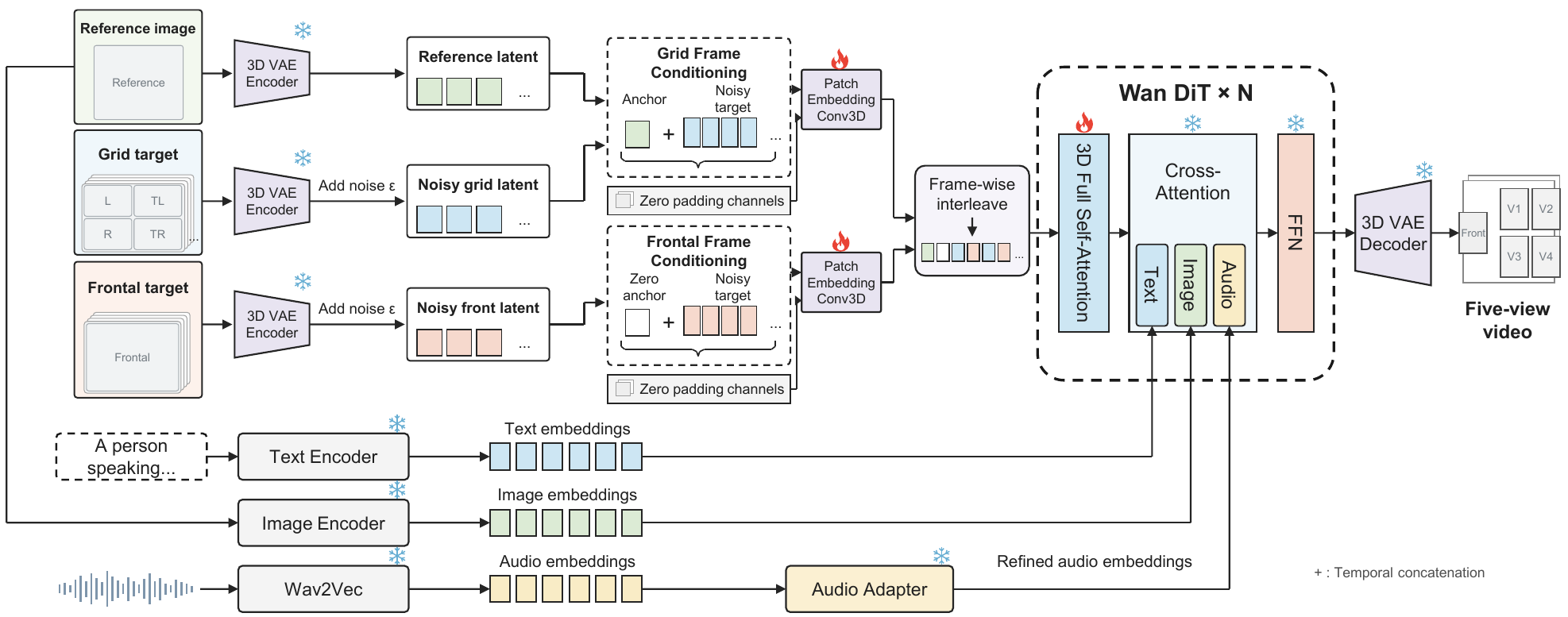}
  \caption{
    \textbf{Stage-I audio-driven multi-view diffusion.}
    A clean reference conditions five jointly denoised views through frame-wise
    token packing and shared self-attention.
  }
  \label{fig:stage1_sparse_tokens}
\end{figure*}

%% file: SEC/A_stage2_implementation.tex
\clearpage
\section{Stage-II Implementation Details}
\label{app:stage2_implementation}

\subsection{Training Camera Pools}
\label{app:stage2_camera_pools}

\paragraph{Camera-holdout variant.}
The qualitative Stage-II component ablations use a 12-camera NeRSemble
training pool.
The four remaining cameras, \texttt{222200046}, \texttt{222200040},
\texttt{222200049}, and \texttt{222200039}, are excluded from both source-view
sampling and real-target supervision throughout training. This preserves a
camera-holdout protocol for assessing generalization beyond the training
viewpoints, separately from the identity-disjoint data split.

\paragraph{Full-camera variant.}
For the main comparisons on RenderMe-360 and HDTF, we retrain from the
released FaceLift checkpoint using the same curriculum and the same
648-clip NeRSemble training subset, but expand the eligible source and
supervision pool to all 16 cameras. The four cameras listed above are
therefore no longer held out for this variant. No RenderMe-360 or HDTF
evaluation data are used for this adaptation.
The quantitative checkpoint comparison in
Table~\ref{tab:stage2_checkpoint_comparison}(b) also uses this 16-camera
training lineage. Its cameras are not held out from training.

\paragraph{Pool size versus input count.}
Both variants reconstruct from five source views; 12 and 16 denote the
training camera-pool sizes, not the number of simultaneous reconstruction
inputs. At inference, both receive the four grid views and the frontal view
generated by Stage~I. The full-camera variant is not used to establish
generalization to the four held-out NeRSemble cameras.

\subsection{View-Aware Corruption}
\label{app:stage2_corruption}

For each sampled set of five source cameras, we keep the most-frontal view
clean and apply corruption to the other four. We identify this anchor by
the smallest angle between a camera's optical axis and that of the frontal
camera \texttt{222200037}, using a fixed ranking derived from the calibrated
NeRSemble rig. The choice is made within the sampled set, preserving the
source-view order. Grid distortion and camera jitter are each enabled with
probability $0.5$, using independent scene-level draws. When enabled, an
operation applies to all four non-anchor views, with independently sampled
warp fields or pose offsets. Both operations can occur in the same sample.
Only source inputs are corrupted; all reconstruction targets and target
cameras remain unchanged.

Grid distortion follows LGM~\citep{tang2024lgm}. We uniformly sample an
integer $n\in\{8,\ldots,16\}$ once per scene and place $n$ control points
along each image axis. For each corrupted view and axis, the normalized
control-point positions are independently perturbed by uniform offsets in
$[-0.25/(n-1),\,0.25/(n-1)]$, corresponding to distortion strength $0.5$.
Positions are clamped to $[0,1]$, converted to pixel coordinates, and the
two endpoints are fixed at the image boundaries. Piecewise-linear
interpolation between control points constructs the sampling grid; bilinear
sampling with \texttt{align\_corners=False} warps the image. Camera
parameters are unchanged by this operation.

Camera jitter samples independent uniform rotation angles in $[-3^\circ,3^\circ]$
about camera-up and $[-1.5^\circ,1.5^\circ]$ about camera-right for each
corrupted view. In the OpenCV camera-to-world convention, these axes are
the negative second column and the first column of the rotation matrix,
respectively. We compose the camera-right rotation followed by the
camera-up rotation and apply the resulting rotation to both the camera
orientation and camera center about the scene origin. This preserves the
camera's distance to the origin while changing its nominal pose; its
intrinsics and source image are unchanged by jitter. The same corruption
settings are used for the 12-camera and 16-camera variants.

\subsection{Active-Volume Regularization}
\label{app:stage2_active_volume}

This section specifies the active-volume loss $\mathcal{L}_{\mathrm{AV}}$
introduced in Section~\ref{sec:method_stage2}. The loss suppresses opacity
outside a canonical head volume and moves opacity-carrying outliers toward
its boundary.

\paragraph{Regularized Gaussians and volume.}
Let $\mathcal P$ index the pixel-aligned Gaussians in a training minibatch,
with centers $\mathbf x_i$ and opacities $a_i$. We apply a sigmoid to the
predicted opacity logits to obtain $a_i$. The two global Gaussians in each
scene are excluded from both the penalty and its normalization. The canonical
head volume is the fixed axis-aligned box
$\mathcal B=[-0.91,0.91]\times[-0.91,0.91]\times[-1,1]$
in the reconstructor's scene coordinates. Writing its lower and upper
corners as $\mathbf l$ and $\mathbf u$, respectively, we compute
\begin{equation}
  \boldsymbol\delta_i
    = \max(\mathbf l-\mathbf x_i,\mathbf 0)
      +\max(\mathbf x_i-\mathbf u,\mathbf 0),
  \qquad
  d_i^2=\|\boldsymbol\delta_i\|_2^2,
  \label{eq:app_stage2_volume_distance}
\end{equation}
where the maxima are element-wise. Thus $d_i$ is the Euclidean distance
from the Gaussian center to the box, and is zero for centers inside or on
its boundary. This definition penalizes center locations rather than the
full spatial extent of each Gaussian.

\paragraph{Loss definition.}
The implementation uses two complementary terms:
\begin{equation}
  \begin{aligned}
    \mathcal{L}_{\mathrm{AV}}
      &= \frac{\sum_{i\in\mathcal P}
                  \mathbf{1}[\mathbf x_i\notin\mathcal B]\,a_i}{Z}
       + \beta\frac{\sum_{i\in\mathcal P}
                  \operatorname{sg}(a_i)\,d_i^2}{Z},\\
    Z &= \max\!\left(\sum_{j\in\mathcal P}
                  \operatorname{sg}(a_j),\,10^{-8}\right).
  \end{aligned}
  \label{eq:stage2_active_volume}
\end{equation}
Here $\mathbf{1}[\cdot]$ is the exterior indicator and
$\operatorname{sg}$ denotes stop-gradient: it preserves a value in the forward
pass but blocks its gradient during backpropagation. The first term is the
fraction of total opacity carried by exterior centers. The second is the
opacity-weighted squared distance to the box. The normalization includes
both interior and exterior pixel-aligned Gaussians. Sums are taken jointly
over the scenes and pixel-aligned Gaussians in each worker's minibatch,
rather than averaging separately normalized per-scene losses.

\paragraph{Gradient handling.}
The first term retains opacity gradients for exterior Gaussians, encouraging
them to become transparent. Its hard exterior indicator does not provide
position gradients. In the distance term, opacity is detached so that this
term provides position gradients rather than an additional gradient for
reducing opacity. These gradients pull active outliers toward the box,
subject to the source-ray constraint of the pixel-aligned representation.
We also detach opacity in $Z$, preventing the normalization from encouraging
increased opacity on interior Gaussians merely to reduce the loss ratio.
The $10^{-8}$ floor prevents division by zero.

\paragraph{Weights.}
We set the relative distance weight to $\beta=1$ and the overall weight to
$\lambda_{\mathrm{AV}}=0.1$ when active-volume regularization is enabled in
Eq.~\eqref{eq:stage2_total_loss}. When it is disabled, its contribution to the
training objective is zero. The volume bounds and distance weight are shared
by the 12-camera and 16-camera training variants.

\begin{figure*}[t]
  \centering
  \includegraphics[width=\textwidth]{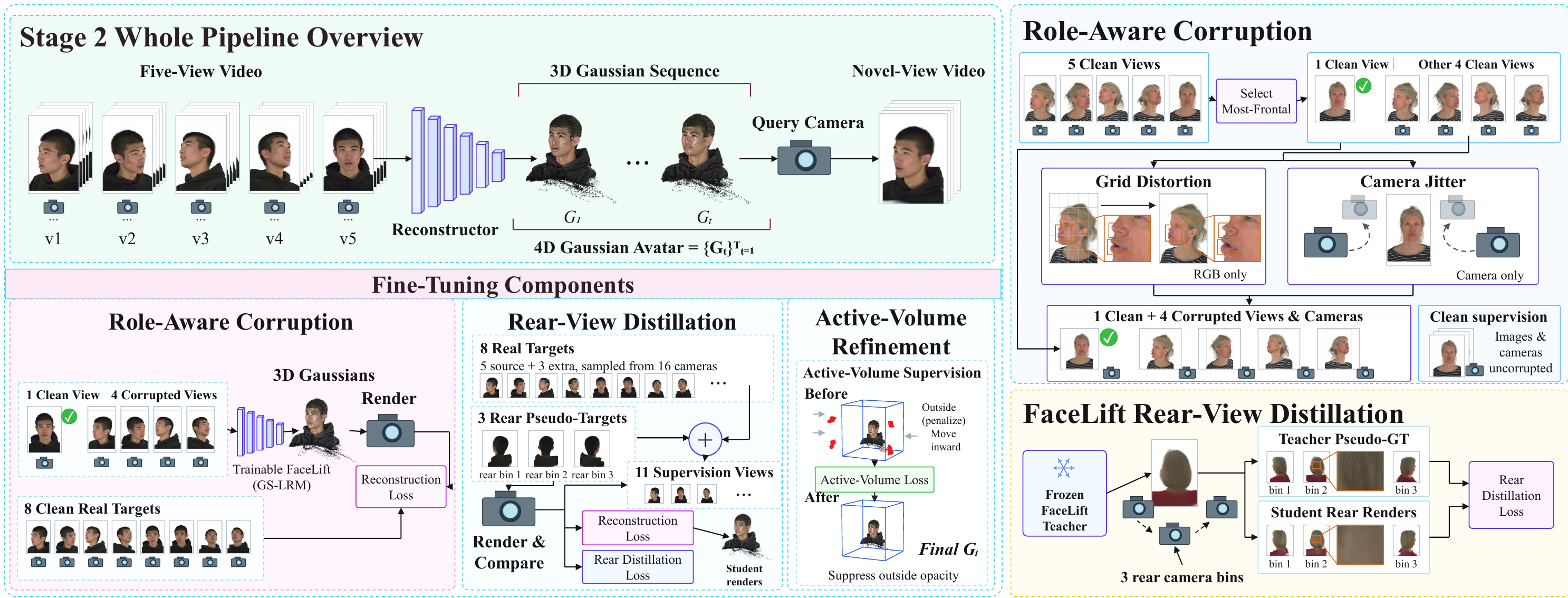}
  \caption{\textbf{Stage-II Gaussian lifting and training curriculum.}}
  \label{fig:stage2_overview}
\end{figure*}

%% file: SEC/A_training_recipe.tex
\clearpage
\section{Training Schedules and Inference Settings}
\label{app:training_recipe}

\paragraph{Stage-I initialization and training.}
We initialize a four-view, clean-reference-conditioned model from the
released 1.3B StableAvatar weights and train transformer self-attention for
9,599 updates, approximately 24 passes over the training data. These trained
four-view weights initialize the five-view model, which undergoes 6,800
additional updates. Table~\ref{tab:stage1_training_schedule} gives the
complete schedule, including the four-view warm start. The
selected cooldown checkpoint is halfway through a 2,000-update linear decay,
at a learning rate of approximately $5\times10^{-5}$. Training uses
81-frame clips, a global batch size of eight for
the five-view runs, seed 42, and AdamW with
$(\beta_1,\beta_2)=(0.9,0.999)$, weight decay $0.03$, $\epsilon=10^{-10}$,
and gradient clipping at $0.05$. Gradient accumulation is one, with
bfloat16 and gradient checkpointing. The trainable modules and exact loss
are specified in Appendix~\ref{app:stage1_training}.

\begin{table}[!htbp]
\centering
\small
\setlength{\tabcolsep}{5pt}
\caption{Stage-I training schedule starting from released StableAvatar weights.
Rows are applied sequentially, and update counts include warmup.}
\label{tab:stage1_training_schedule}
\begin{tabular}{lrrrl}
\toprule
Phase & Updates & Nominal batch & Warmup & Learning-rate schedule \\
\midrule
Four-view warm start & 9,599 & 8 & 500 & Constant $10^{-4}$ after warmup \\
\midrule
Five-view adaptation & 4,200 & 8 & 400 & Constant $5\times10^{-4}$ after warmup \\
Continuation & 1,600 & 8 & 50 & Constant $10^{-4}$ after warmup \\
Cooldown & 1,000 & 8 & 0 & Linear $10^{-4}\to0$ over 2,000 updates \\
\bottomrule
\end{tabular}
\end{table}

\paragraph{Stage-II data and training.}
The two camera-pool variants in Appendix~\ref{app:stage2_camera_pools}
follow the same schedule in Table~\ref{tab:stage2_training_schedule},
starting from released FaceLift. The 648 training clips supply 16 frames
each, at indices $0,5,11,16,21,27,32,37,43,48,53,59,64,69,75,80$,
giving 10,368 scenes. A fixed seed-103 camera-sampling schedule contains five
exposures per scene, or 51,840 training examples, with five source views and
three additional real target views; all eight views receive reconstruction
supervision. The two 3,240-update adaptation phases each traverse this
schedule once. All source images are captured observations. Source
backgrounds remain white, while a background sampled uniformly from white,
black, and mid-gray (RGB $0.5$) is shared by the targets and their renders, but is not
provided to the transformer. All five phases use AdamW with
$(\beta_1,\beta_2)=(0.9,0.95)$, weight decay $0.05$ on matrix parameters
(zero on biases and normalization vectors), gradient clipping at 1,
bfloat16, gradient accumulation of one, and gradient checkpointing.
Each phase initializes fresh optimizer and scheduler state from the selected
preceding weights and uses 20 linear warmup updates followed by cosine
decay. The full-camera model after the final 150-update refinement is used
for the external comparisons.

\begin{table}[!htbp]
\centering
\small
\setlength{\tabcolsep}{3pt}
\caption{Stage-II training schedule, totaling 7,328 selected updates.
Horizon denotes the total cosine-scheduler length. Each phase has 20 warmup
updates. View-aware corruption remains enabled after its introduction.}
\label{tab:stage2_training_schedule}
\begin{tabular}{lrrrccc}
\toprule
Phase & Updates & Batch & Peak LR & Horizon
      & $\lambda_{\mathrm{rear}}$ & $\lambda_{\mathrm{AV}}$ \\
\midrule
Camera/background adaptation & 3,240 & 16 & $8\times10^{-5}$ & 3,240 & 0 & 0 \\
View-aware corruption & 3,240 & 16 & $8\times10^{-5}$ & 3,240 & 0 & 0 \\
Active-volume refinement & 50 & 12 & $10^{-5}$ & 864 & 0 & 0.1 \\
Rear-prior distillation & 648 & 16 & $10^{-5}$ & 3,240 & 0.2 & 0 \\
Joint rear/active-volume refinement & 150 & 16 & $10^{-5}$ & 3,240 & 0.2 & 0.1 \\
\bottomrule
\end{tabular}
\end{table}

\paragraph{Corruption seeds and rear targets.}
Source-view selection and corruption operators are specified in
Appendix~\ref{app:stage2_corruption}. Corruption uses seed 84, except seed 8403
in the 50-update active-volume phase; background sampling uses seed 100.
The trainer separately seeds PyTorch with $777+\mathrm{rank}$.
The frozen teacher cache is built from each scene's fixed five clean
canonical views (cameras \texttt{222200042}, \texttt{221501007},
\texttt{222200044}, \texttt{222200045}, and \texttt{222200037}), not the
student's randomly sampled tuple. With camera seed 105, one rear view is
sampled from each of the yaw intervals $[110^\circ,157^\circ)$,
$[157^\circ,203^\circ)$, and $[203^\circ,250^\circ)$, and the three cached
teacher targets are composited over the same sampled background as the real
targets. Real and teacher rendering losses both use pixel MSE plus
$0.5$ times the inherited VGG19 perceptual loss; the latter is distinct
from the AlexNet LPIPS evaluation metric. Teacher supervision is weighted
by $0.2$ in both final phases. Active-volume settings are given in
Appendix~\ref{app:stage2_active_volume}, with weight $0.1$ in the two
indicated refinement phases.

\paragraph{Inference and long sequences.}
Stage I uses 30 flow-matching Euler steps, scheduler shift 5 with dynamic
shifting disabled, seed 42, and bfloat16 inference. Audio and text guidance
scales are 5 and 3, respectively, with the prompt ``The protagonist is
talking'' and StableAvatar's default negative prompt.
The four-view canvas is $832\times1216$ and the frontal stream is
$416\times608$ (width $\times$ height); splitting the canvas therefore gives
five native-resolution $416\times608$ views at 25 fps. Longer sequences use
81-frame windows, corresponding to 21 latent frames, with a nominal overlap
of five \emph{latent} frames and a stride of 16. At every denoising step,
both streams process the same windows and linearly crossfade overlapping
latent updates; the first denoising step skips crossfading. The final window
is shifted back to remain full-length when necessary, increasing its overlap.
Stage II
then reconstructs each time index independently from the five calibrated
views using float16 autocast. All requested views at a given time render the
same full set of predicted Gaussians.
The geometry-refinement configuration retains its log-scale cap of $-1.20$;
external-evaluation renders are $832\times1216$ before the common-canvas
preprocessing in Appendix~\ref{app:metric_frontal_shared}.

%% file: SEC/A_dataset_baselines.tex
\clearpage
\section{Dataset Construction and Baseline Generation}
\label{app:dataset_baselines}

\paragraph{Training data and identity split.}
We use the speech (SEN) recordings of NeRSemble. Data preparation retains
participants with at least two SEN recordings and requires the designated
source-camera videos, background reference images, and audio. Incomplete
camera data and clips with missing audio are excluded; recordings must be
long enough to supply 81 frames at 25 fps. For each retained recording, we
take the first 81-frame window, sampling raw frame
$\operatorname{round}(73t/25)$ for output index $t=0,\ldots,80$, together
with the corresponding 3.24-second, 16-kHz mono audio segment.
The final split contains 2,641 training clips from 328 identities and
81 development clips from ten different identities: 067, 145, 177, 222,
279, 302, 369, 413, 476, and 490. All clips of an identity belong to the
same split. These ten identities are used for development and checkpoint
selection. The eight-person quantitative
ablation cohort in Appendix~\ref{app:quantitative_ablation_evaluation} is a subset of
this development split. RenderMe-360 and HDTF are external evaluations;
neither is used to train or fine-tune our two stages.

\paragraph{Image preparation and reconstruction subset.}
We remove NeRSemble backgrounds with BackgroundMattingV2 using the
ResNet-101 checkpoint, backbone scale $0.25$, sampling refinement, and
80,000 refinement pixels. Each source frame and its recorded empty
background are processed together, and the predicted foreground and alpha
are composited over white. Each view is resized with Pillow Lanczos to fit
within $416\times608$ while preserving aspect ratio, then centered with
white padding; mask padding is zero. The four grid cells use cameras
\texttt{222200042}, \texttt{221501007}, \texttt{222200044}, and
\texttt{222200045} in top-left, top-right, bottom-left, and bottom-right
order. Camera \texttt{222200037} supplies the separate frontal stream and
the reference portrait. Stage II uses a fixed 648-clip subset spanning all
328 training identities, with approximately two clips per identity to
balance their contributions. It uses captured synchronized images. The 16
sampled time indices per clip, repeated camera-tuple exposures, and target
sampling are specified in Appendix~\ref{app:training_recipe}.
The 12-camera variant excludes \texttt{222200046}, \texttt{222200040},
\texttt{222200049}, and \texttt{222200039} from both source sampling and
real-target supervision. The 16-camera variant includes these cameras and
uses the same identity split and 648 clips. Missing-camera scenes
sample only cameras actually available. Both variants take five source views.

\subsection{Baseline Video Generation}
\label{app:baseline_generation}

The following settings apply on both RenderMe-360 and HDTF. Each baseline
is conditioned only on the same underlying reference portrait and
51,840-sample speech segment as our method. Its input square is the crop
$(0,191,832,1023)$ of the frozen
$832\times1216$ portrait, stored as an $832\times832$ JPEG at quality 95.
Method-specific face processing is applied to this square, as described
below. All outputs have 81 frames at 25 fps. The current evaluation cohorts,
GT construction, output-to-scoring transformations, and aggregation are
specified in Appendix~\ref{app:metric_psnr_ssim}.

\paragraph{Real3D-Portrait.}
We use the released \path{240210_real3dportrait_orig} models:
\path{audio2secc_vae} at checkpoint 400,000 and
\path{secc2plane_torso_orig} at checkpoint 100,000. Inference uses
seed 42, static source pose (\texttt{drv\_pose=static}),
\texttt{out\_mode=final}, periodic blinking, audio-to-motion temperature
$0.2$, and mouth amplitude $0.45$. We retain the released source-face crop
threshold $0.2$, initial-pose mapping, and head/torso/background compositing.
The portrait supplies the static pose and background. The source
is processed at $512\times512$ by the released face-cropping and 3DMM
estimation path. Audio is processed by the released HuBERT/F0 path, including
its motion-padding convention. The internally padded 84-frame rendering is
muxed with the 3.24-second audio by the original shortest-stream export,
yielding the evaluated 81-frame $512\times512$ video.

\paragraph{SDTalk.}
We use \texttt{SDGTalk.pt} and its stored model configuration, with seed 42
for each clip. The source square is bilinearly resized to $512\times512$
with antialiasing, segmented with the released
\texttt{jonathandinu/face-parsing} SegFormer, and masked to labels 1--18
with black elsewhere. The released source tracker estimates FLAME shape,
expression, jaw/eye codes, and the source camera. For source eye landmarks,
we select the S3FD detection with the highest confidence, breaking exact
ties by bounding-box area. The source camera and six eye measurements
remain fixed throughout the sequence; only audio changes over time.
We retain the checkpoint's FLAME scale 5, native focal parameter 12,
$296\times296$ source lifting planes, $518\times518$ image-encoder input,
and released feature-to-RGB decoder. PCM16 audio is divided by 32,768 and
normalized as $(x-\operatorname{mean}(x))/\sqrt{\operatorname{var}(x)+10^{-7}}$.
The last hidden states of \texttt{facebook/hubert-large-ls960-ft} give
161 vectors of dimension 1,024 for the 3.24-second segment. Video frame $t$
uses the 16-vector window with indices $2t-8,\ldots,2t+7$, with out-of-range
indices filled by zero vectors. This produces an $81\times16\times1024$
audio tensor. The model outputs $512\times512$ RGB. Prediction-only
white-background conversion is applied during evaluation.

\paragraph{NeRFFaceSpeech.}
We use the released real-image/audio pipeline with
\path{ffhq_1024.pkl}, \texttt{LipaintNet.pt}, and the audio-to-expression
weights from \path{SadTalker_V0.0.2_256.safetensors}. The input square is
Lanczos-resized to $1024\times1024$ and passed through the released PTI/dlib
face alignment using the 68-landmark predictor. We consistently select the
largest detected face by inclusive bounding-box area, choosing the first
detection on ties. The selected landmarks are passed to the released
alignment, retaining its crop, padding, and bilinear quadrilateral transform
to $1024\times1024$. Each clip starts a fresh source-image inversion and
pivotal tuning. Inversion
runs 1,000 updates, jointly optimizing the latent, noise, and background
codes with Adam $(0.9,0.999)$, a peak learning rate of $0.01$, a linear
5\% ramp-up and cosine 25\% ramp-down. The latent mean uses 600 samples
with seed 123; initial latent-noise strength is $0.05$, decaying to zero
over 75\% of the updates. The inversion objective is the released VGG16
feature distance plus noise regularization weighted by $10^5$. PTI uses
Adam at $3\times10^{-4}$ for at most 2,000 updates, stopping when the source
AlexNet LPIPS reaches $0.06$. Its pixel-MSE and LPIPS weights are both one,
and locality regularization is disabled. The subsequent released BFM09
fit uses a $224\times224$ image, 1,000 rigid and 500 non-rigid updates,
both at learning rate $0.01$.
Generation uses seed 0, truncation $0.7$, constant synthesis noise, the
source-estimated camera, frozen background, expression scale $1.5$, and
the released $0.5$ reference-expression update weight. We retain the
seven-frame moving average of the mouth-inpainting mask and the released
deformation/inpainting operations. Native output is $1024\times1024$;
benchmark export retains the generated frames before the presentation
watermark.

\paragraph{Video export and scoring input.}
Frontal metrics decode each method's exported videos. Real3D-Portrait
retains its released imageio H.264 export and FFmpeg audio mux. SDTalk uses
H.264 CRF 15 and NeRFFaceSpeech uses H.264 CRF 18, both with
\texttt{yuv420p} pixels at 25 fps. We verify that each exported video
contains all 81 frames. Frontal appearance scoring then follows
Appendix~\ref{app:metric_frontal_shared}; the identity protocol remains the
separate native-video procedure in Appendix~\ref{app:metric_csim_ref_v8}.

\subsection{Calibrated Novel-View Rendering}
\label{app:baseline_cameras}

All five scored target cameras render the same frame-0 identity/expression
state for a clip. We use physical RenderMe-360 camera
calibration and the source frame's 73 three-dimensional face landmarks to
register coordinate systems. These metadata are used only for camera
registration. Each clip has one similarity mapping
$X'=sQX+b$. A target camera with camera-to-world rotation $R_j$ and center
$c_j$ is transferred as $R'_j=QR_j$ and $c'_j=sQc_j+b$; its intrinsics
follow the same frozen crop/resize used for that camera's GT. The same
mapping is shared across all target cameras of a clip.

\paragraph{Real3D-Portrait camera adapter.}
The native camera is the source-fitted EG3D camera. We fix the canonical
head anchor to $(0,0,0.45)$, disable mirroring, and obtain the initial scale
from the ratio of native-camera-to-anchor distance to
physical-source-camera-to-landmark-centroid distance. These method-level
choices were calibrated on RenderMe-360 subject 0041 using target-view
diagnostics, checked on subject 0290, and then held fixed for all clips. With
scale and source-camera rotation fixed, four source-framing iterations
adjust translation, first along the optical axis to match the source
landmarks' RMS image radius, then laterally to match their image centroid.
The resulting cameras are used directly for rendering.
Target intrinsics are passed to the EG3D ray sampler with their first two
rows divided by 512. Identity, audio-frame-0 expression, and the released
head/torso compositor remain unchanged across targets.

\paragraph{SDTalk camera adapter.}
We retain the frame-0 audio-deformed Gaussian cloud and source lifting
planes. The physical source pose is estimated with OpenCV iterative PnP
from source FLAME/FAN landmarks 17--67 (zero-based), the source intrinsics,
and the native pose initialization.
Scale is the ratio of the FLAME and physical-source interocular distances:
FLAME eye centers use indices 36--41 and 42--47, and the RenderMe-360
73-point annotation uses indices 19--24 and 25--30. The physical and fitted
source cameras then determine $Q$ and $b$. The target camera rasterizes
the same 32-channel Gaussian features, after which the released StyleUNet
produces RGB. Off-axis intrinsics are implemented in the projection and
feature-raster canvas, cropped to $512\times512$ before decoding.
The decoder noise state is shared across
target views. The fixed prediction-only background conversion is applied
after rendering, as in the frontal evaluation.

\paragraph{NeRFFaceSpeech camera adapter.}
We reuse each clip's source-image fit and audio-frame-0 expression state.
The known source crop, resize, and PTI alignment are composed into an image
transform; RQ factorization of its product with the physical source
intrinsics accounts for alignment roll. The canonical head anchor is the
origin, and the initial scale is the native-source-camera distance divided
by the physical-source-camera-to-landmark-centroid distance. Four
source-framing depth/lateral iterations, as above, set translation while
keeping scale and rotation fixed. Each transferred camera uses the full
inverse-NDC intrinsic matrix and OpenGL camera-to-world matrix, with the
proper basis change $\operatorname{diag}(1,-1,-1)$ from OpenCV coordinates.
For $512\to1024$ rendering, focal lengths double and principal points
transform as $c'=2c+0.5$. Camera-dependent mouth inpainting is recomputed
at each view; the released ray-slot deformation and frozen-background
behavior are retained.
Native $1024\times1024$ novel-view renders are Lanczos-downsampled to
$512\times512$ and scored directly.